\documentclass[conference]{IEEEtran}
\IEEEoverridecommandlockouts
\usepackage{cite}
\usepackage{amsmath,amssymb,amsfonts}
\usepackage{amsthm}
\usepackage{algorithmic}
\usepackage{algorithm}
\usepackage{graphicx}
\usepackage{textcomp}
\usepackage{pdfpages}
\usepackage[bookmarks=false]{hyperref}

\usepackage{subfig}
\usepackage[english]{babel}
\makeatother
\theoremstyle{plain}
\newtheorem{lem}{Lemma}
\theoremstyle{plain}
\newtheorem{thm}[lem]{Theorem}
\theoremstyle{plain}
\newtheorem{prop}[lem]{Proposition}

\global\long\def\gv{\given}
\global\long\def\given{\mid}

\global\long\def\argmin#1{\underset{_{#1}}{\text{argmin}\ } }    
\begin{document}

\title{Accelerating Experimental Design by Incorporating Experimenter Hunches}

\author{\IEEEauthorblockN{Cheng Li\IEEEauthorrefmark{1}, Santu Rana\IEEEauthorrefmark{1}, Sunil Gupta\IEEEauthorrefmark{1}, Vu Nguyen\IEEEauthorrefmark{1}, Svetha Venkatesh\IEEEauthorrefmark{1}, \\Alessandra
Sutti \IEEEauthorrefmark{2}, David Rubin\IEEEauthorrefmark{2}, Teo Slezak\IEEEauthorrefmark{2},  Murray Height\IEEEauthorrefmark{3}, Mazher Mohammed\IEEEauthorrefmark{4}, and Ian Gibson\IEEEauthorrefmark{4}}
\IEEEauthorblockA{\IEEEauthorrefmark{1} Deakin University, Geelong, Australia, PRaDA \\
\IEEEauthorrefmark{2} Deakin University, Geelong, Australia, IFM. \IEEEauthorrefmark{3} HeiQ Australia, Pty Ltd \\ \IEEEauthorrefmark{4}Deakin University, Geelong, Australia, School of Engineering}}

\maketitle

\begin{abstract}
Experimental design is a process of obtaining a product with target
property via experimentation. Bayesian optimization offers a sample-efficient
tool for experimental design when experiments are expensive. Often,
expert experimenters have 'hunches' about the behavior of the experimental
system, offering potentials to further improve the efficiency. In
this paper, we consider per-variable monotonic trend in the underlying
property that results in a unimodal trend in those variables for a
target value optimization. For example, sweetness of a candy is monotonic
to the sugar content. However, to obtain a target sweetness, the utility
of the sugar content becomes a unimodal function, which peaks at the
value giving the target sweetness and falls off both ways. In this
paper, we propose a novel method to solve such problems that achieves
two main objectives: a) the monotonicity information is used to the
fullest extent possible, whilst ensuring that b) the convergence guarantee
remains intact. This is achieved by a two-stage Gaussian process modeling,
where the first stage uses the monotonicity trend to model the underlying
property, and the second stage uses `virtual' samples, sampled from
the first, to model the target value optimization function. The process
is made theoretically consistent by adding appropriate adjustment
factor in the posterior computation, necessitated because of using
the `virtual' samples. The proposed method is evaluated through both
simulations and real world experimental design problems of a) new
short polymer fiber with the target length, and b) designing of a
new three dimensional porous scaffolding with a target porosity. In
all scenarios our method demonstrates faster convergence than the
basic Bayesian optimization approach not using such `hunches'.
\end{abstract}

\IEEEpeerreviewmaketitle
\section{Introduction}

Experimental design involves optimizing towards a target goal by iteratively
modifying often large numbers of control variables and observing the
result. For hundreds of years, this method has underpinned the discovery,
development and improvement of almost everything around us. When experimental
design entails an expensive system then Bayesian optimization \cite{brochu_tutorial_2010}
offers a sample-efficient method for global optimization. Bayesian
optimization is a sequential, model-based optimization algorithm,
which uses a probabilistic model, often a Gaussian process, as a posterior
distribution over the function space. Based on the probabilistic model
an utility function is constructed to seek the best location to sample
next, such that the convergence towards global optima happens quickly
\cite{srinivas10gaussian}. The detail of Bayesian optimization is
provided in background section \ref{subsec:Bayesian-Optimization}.
It has been used in many real world design problems including alloy
design \cite{Balachandran2016AdaptiveSF,Vu_Rana_Gupta_Li_Venkatesh_ICDM16}, short polymer fiber design
\cite{Li_rapidBO_2017}, and more commonly, in machine learning hyper-parameter
tuning \cite{feurer2015initializing,Li_BO_dropout_2017,pmlr-v70-rana17a}. However, a generic Bayesian
optimization algorithm is under-equipped to harness intuitions or
prior knowledge, which may be available from expert experimenters.
\begin{figure}
\begin{centering}
\includegraphics[width=0.95\columnwidth]{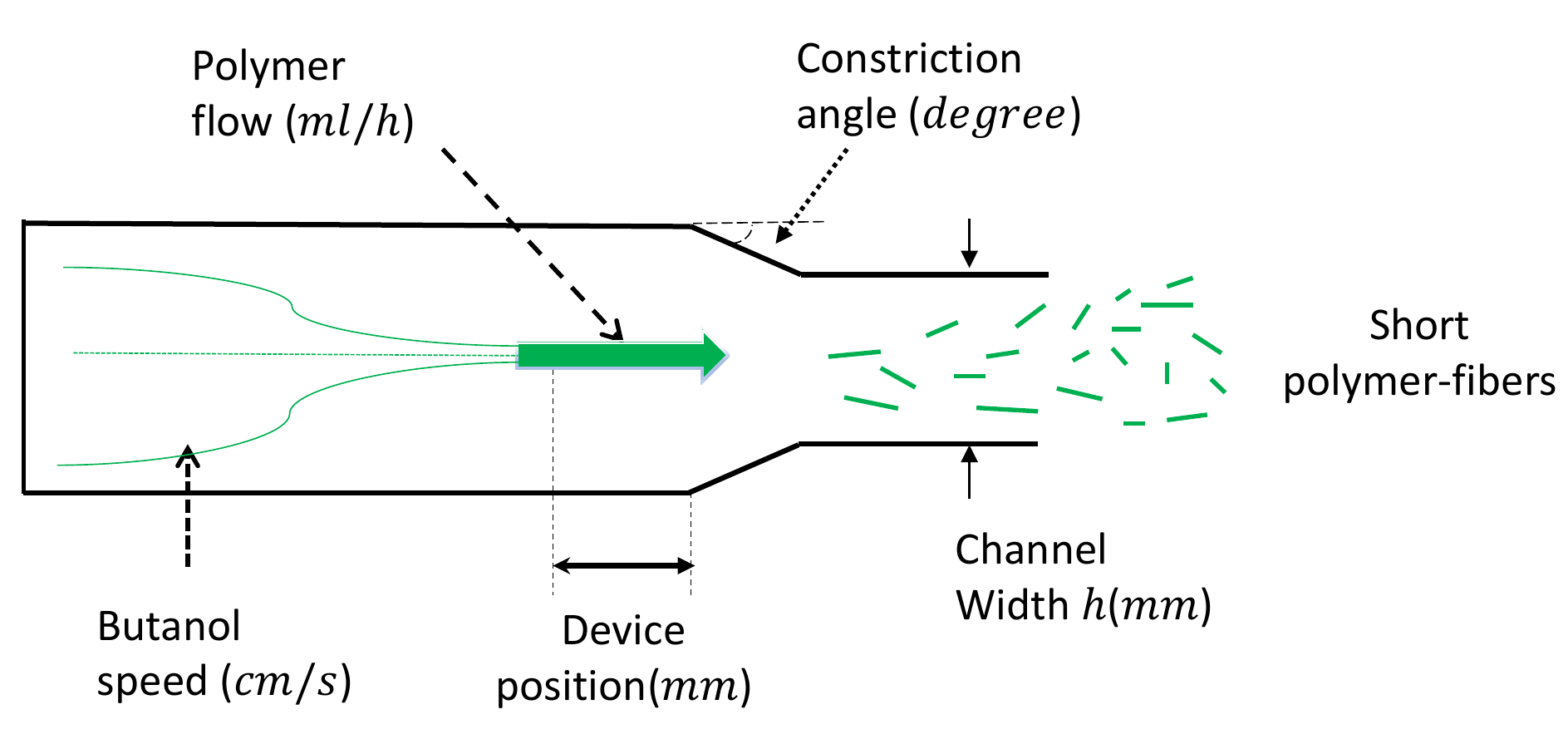}
\par\end{centering}
\caption{\label{fig:SPF_flowchart}Short polymer fiber (SPF) synthesis using
a microfluidic device. This device is parameterized by five parameters:
\emph{geometric factors}: channel width ($mm$), constriction angle
($degree$), and device position ($mm$); and,\emph{ flow factors}:
coagulant (e.g. butanol) speed ($cm/s$), polymer flow ($ml/h$). }
\end{figure}

Consider the production of short polymer fibers with specific length
and diameter as an experimental design problem. These fibers are used
to coat natural fabrics to make them superior in many aspects e.g.
more resistive to pilling, improved water repellence etc. Different
types of fabrics generally require different sizes of the fibers for
optimal results. The fibers are produced by injecting a polymer liquid
through a high speed coagulant (e.g. butanol) flow inside a specially
designed apparatus (see Figure \ref{fig:SPF_flowchart}). The differential
speed between the polymer and the coagulant flows turns the liquid
polymer into short and thin nano-scale fibers. The geometrical parameters
of the apparatus and the flow speeds control the shapes and sizes
of the fibers produced. In order to produce fibers with the specific
length and diameter, we need to find the right values for these control
parameters. Since the whole process of producing fibers is expensive,
we expect to achieve the desired product by using fewer experiments.
Bayesian optimization offers a perfect choice for this task. However,
in this fiber production, experimenters have a prior knowledge that
fiber length monotonically decreases with respect to the coagulant
flow speed. Such `hunches' can be directly useful in cutting down
the search space if one is interested in producing either the shortest
or the longest fibers. But they are not straightforwardly useful for
our problem of producing fibers with a target length. In this case,
such hunches do not reduce the search space, but they could still
be useful in reducing the model space for model-based optimization
algorithms, such as Gaussian process (GP) in Bayesian optimization.
With a smaller model space to search from, it might be possible that
the convergence of optimizer happens quicker. 

Formally, our optimization problem based on a target $y_{T}$ can
be written as,
\begin{equation}
\boldsymbol{x}^{*}=\text{argmin}_{\boldsymbol{x}\in\mathcal{X}}g(\boldsymbol{x})\triangleq|f(\boldsymbol{x})-y_{T}|\label{eq:objective function}
\end{equation}
where $f(\boldsymbol{x})$ maps control variables $\boldsymbol{x}$
to the measured property. For example, in the already mentioned polymer
fibre design problem, $\boldsymbol{x}$ is a vector of five parameters
shown in Figure \ref{fig:SPF_flowchart}, $f$ is the measured fiber
length and $y_{T}$ is a target length. The hunch that the experimenters
posses is that fiber length is monotonically decreasing with the coagulant
flow. Whilst the resultant function $g(\boldsymbol{x})$ is still
a complex function over all the variables, but across the coagulant
flow it is guaranteed to be unimodal. When performing Bayesian optimization,
such knowledge can be useful in building a more accurate posterior
Gaussian process of $g(\boldsymbol{x})$. In our experience we have
found that humans are more comfortable in giving per-variable trends
than multivariable ones. Also, hunches about monotonicity are more
available than any more complex trends. Thus, in this work, we only
consider hunches which are simple per variable monotonicity trends,
which results in a target value optimization function that is unimodal
in those variables. 

Some of the recent work has examined various mechanisms to incorporate
prior shape information into GP modeling including the enforcement
on monotonicity \cite{pmlr-v9-riihimaki10a,Wang_SIAM_2016} and monotone-convex/concavity
\cite{CIS-472054}. Wu et al. \cite{Wu_NIPS2017_7111} has considered
incorporation of exact derivative values in Bayesian optimization,
but exact derivatives are hard to acquire in practice. Preliminary
work \cite{Jauch_nips_shapeBO_2016,Andersen_unimodal_2017} enforces
unimodality by controlling derivative sign. Unfortunately, Jauch and
Pena \cite{Jauch_nips_shapeBO_2016} requires specification of the
turning point, thus severely restricting application of their algorithms.
Andersen et al. \cite{Andersen_unimodal_2017} needs to compute an
intractable marginal from a complex joint distribution. Surprisingly,
there is no details of the inference process in \cite{Andersen_unimodal_2017}
and thus we were unable to verify or replicate their approach. Hence,
we can safely conclude that none of the existing works in Bayesian
optimization solves our problem where objective function is unimodal
in certain dimensions, thus the problem remains open.

Our approach is based on correctly converting the monotonicity information
of $f(\boldsymbol{x})$ to the unimodality information of $g(\boldsymbol{x})$
and then building a better Gaussian process model for $g(\boldsymbol{x})$.
This is non-trivial since monotonicity implies a fixed sign for derivative
of $f(\boldsymbol{x})$, whereas unimodality implies reversal in the
sign of derivatives for $g(\boldsymbol{x})$ at the turning point.
For our case we do not know the location of the turning point. In
absence of turning point, a naive way can be used to derive derivative
signs for $g(\boldsymbol{x})$ based on current knowledge. Specifically,
based on the monotonicity direction and whether $f(\boldsymbol{x})$
is greater or smaller than the target ($y_{T}$), we can appropriately
give +1 or -1 signs on some locations of $g(\boldsymbol{x})$. For
example, for a minimization problem if $f(\boldsymbol{x})$ is monotonic
with decreasing direction then we can put -1 at the locations where
$f(\boldsymbol{x})>f_{T}$ and +1, otherwise. A more information rich
GP model for $g(\boldsymbol{x})$ can be then built by combining the
derived derivative signs and the available observation set $\{\boldsymbol{x},g(\boldsymbol{x})\}$
using the framework of \cite{pmlr-v9-riihimaki10a}. Although this
na\"ive idea is consistent, we show that this leads to severe under-utilization
of the monotonicity information. As shown in Figure \ref{fig:posteriorGP}(b),
a vast region may remain ambiguous to which sign the derivative of
$g(\boldsymbol{x})$ should take. 

Hence, our proposed approach is built in a two-stage process to achieve
two important objectives, a) maximally use the monotonicity information,
leaving no ambiguous region and b) theoretically remain consistent.
We first model $f(\boldsymbol{x})$ through a Gaussian process ensuring
that the mean function is monotonic in the desired variables. We then
sample ``virtual observations'' from the posterior GP of $f(\boldsymbol{x})$
and combine them with real observations to model $g(\boldsymbol{x})$
through another Gaussian process. Since we can sample virtual observation
wherever we want, we do not face the problem of having ambiguous regions
again (Figure \ref{fig:posteriorGP}(d)). However, this may lead to
theoretical inconsistency. The reason is, the GP model of $g(\boldsymbol{x})$
using those virtual observations not only can fix the mean function,
but also may reduce the epistemic uncertainty of $g(\boldsymbol{x})$
by an equal measure. While the former is desirable, too much of the
latter is undesirable, since the correct computation of epistemic
uncertainty is critical for the success of Bayesian optimization \cite{desautels14parallelizing}.
To fix this, we theoretically derive an adjustment factor which corrects
the overconfidence and ensures that our approach remains consistent.

We first demonstrate our methods on synthetic functions and hyperparameter
tuning of neural networks. Then we solve two real world experimental
design problems: a) design of short-polymer fibers with specific length,
and b) design of 3d printed scaffolding with a target porosity. We
use monotonicity information available from the experimenters. We
demonstrate that our method outperforms the generic Bayesian optimization
in these complex experimental design tasks in terms of reduced number
of experimentation to reach target, saving both cost and time. The
significance lies in the fact that such 'hunches' are widely available
from experimenters from almost every domain, and thus the ability
of using them to accelerate experimental design process will further
boost a wider adoption of Bayesian optimization in real world product
and process design.

\section{Background\label{sec:bg}}

\subsection{\label{subsec:GP-Signs}Gaussian Process with Derivative Signs}

Let \textbf{$\boldsymbol{x}$} be a random $D$-dimensional vector
in a compact set $\mathcal{X}:$ $\mathcal{X\rightarrow\mathbb{R}}$.
We denote $\mathcal{D}=\{\boldsymbol{x}_{i},y_{i}\}_{i=1}^{t}$ as
a set of observations, where $y_{i}=f(\boldsymbol{x}_{i})+\varepsilon_{i}$
is the noisy observation of $f(\boldsymbol{x})$ at $\boldsymbol{x}_{i}$
and $\varepsilon_{i}\sim\mathcal{N}(0,\sigma_{noise}^{2})$. A Gaussian
process (GP) \cite{Rasmussen:2005:GPM:1162254} is a random process
such that every finite subset of variables has a multivariate normal
distribution. A GP prior on a latent objective function $f(\boldsymbol{x})$
is fully specified by its mean function $\mu(\boldsymbol{x})=\mathbb{E}[f(\boldsymbol{x})]$
and the covariance function $k(\boldsymbol{x},\boldsymbol{x}^{'})=\mathbb{E}[(f(\boldsymbol{x})-\mu(\boldsymbol{x}))(f(\boldsymbol{x}^{'})-\mu(\boldsymbol{x}^{'}))]$.
A zero-mean GP prior is formulated as
\begin{equation}
f(\boldsymbol{x})\sim\mathcal{GP}(\boldsymbol{0},k(\boldsymbol{x},\boldsymbol{x}^{'}))
\end{equation}
The kernel function $k$ encodes the prior belief regarding the smoothness
of the objective function. A popular kernel is the square exponential
(SE) function $k(\boldsymbol{x}_{i},\boldsymbol{x}_{j})=\epsilon\exp(-\frac{1}{2l^{2}}||\boldsymbol{x}_{i}-\boldsymbol{x}_{j}||^{2})$,
where $\epsilon$ is the output variance and $l$ is the length scale.
The predictive distribution of $y^{+}$ for a test point $\boldsymbol{x}^{+}$
in GP can be computed by\texttt{\small{}
\begin{equation}
y^{+}\gv\boldsymbol{y}_{1:t}\sim\mathcal{N}(\mathbf{k}^{T}\mathbf{K}^{-1}\boldsymbol{y}_{1:t},k(\boldsymbol{x}_{t+1},\boldsymbol{x}_{t+1})-\mathbf{k}^{T}\mathbf{K}^{-1}\mathbf{k})\label{eq:predictive distribution-2}
\end{equation}
}where $\mathcal{N}$ denotes a Gaussian distribution, $\mathbf{k}=[k(\boldsymbol{x}^{+},\boldsymbol{x}_{1})\,\cdots\,k(\boldsymbol{x}^{+},\boldsymbol{x}_{t})]^{T}$
and $\mathbf{K}=[k(x_{i},x_{j})]_{i,j\in\{1,\cdots,t\}}+\sigma_{noise}^{2}\mathbf{I}$.

Since the GP is a linear operator, the derivative of Gaussian process
is still a Gaussian process \cite{pmlr-v9-riihimaki10a}. Therefore,
incorporating derivative values into GP for prediction is straightforward
since the joint distribution of derivative value and function value
is still a Gaussian distribution. In our work it is hard to acquire
derivative values and we only have derivative signs derived from the
prior monotonicity knowledge. The derivative sign '+1' denotes that
the gradient of latent function at the location is positive and '-1'
denotes that the gradient is negative at this location. We follow
the work in \cite{pmlr-v9-riihimaki10a} to compute the posterior
GP given function observations and derivative signs. 

Let $\mathcal{M}=\{\boldsymbol{x}_{s_{i}},s_{i}\}_{i=1}^{m}$ denote
$m$ derivative sign observations, where $s_{i}$ is the derivative
sign at location $\boldsymbol{x}_{s_{i}}$. We specify the derivative
sign as the partial one with respect to the $d$th variable. It is
also easy to extend to any number of variables. For convenience, we
denote $X=\{\boldsymbol{x}_{i}\}_{i=1}^{t}$, $X_{s}=\{\boldsymbol{x}_{s_{i}}\}_{i=1}^{m}$
and $\boldsymbol{s}=\{s_{i}\}_{i=1}^{m}$. The latent function value
and the partial derivative value for the $d$th variable are denoted
as $\boldsymbol{f}$ and $\boldsymbol{f}^{'}$ respectively. 

In Gaussian process regression, the goal is to compute the posterior
predictive distribution of a test point. Similarly, given observations
and derivative signs we can express the predictive distribution of
a test point $\boldsymbol{x}^{+}$ by integrating out the latent $\boldsymbol{f}$
and $\boldsymbol{f}^{'}$
\begin{align}
 & p(y^{+}\gv\boldsymbol{x}^{+},X,\boldsymbol{y},X_{s},\boldsymbol{s})=\nonumber \\
 & \int p(y^{+}\gv\boldsymbol{x}^{+},X,\boldsymbol{y},\boldsymbol{f,}X_{s},\boldsymbol{s},\boldsymbol{f}^{'})p(\boldsymbol{f},\boldsymbol{f}^{'}\gv X,\boldsymbol{y},X_{s},\boldsymbol{s})d\boldsymbol{f}d\boldsymbol{f}^{'}\label{eq:predictive}
\end{align}
The first term $p(y+\gv\boldsymbol{x}^{+},X,\boldsymbol{y},\boldsymbol{f},X_{s},\boldsymbol{s},\boldsymbol{f}^{'})$
at the right side above is a Gaussian distribution (see \cite{pmlr-v9-riihimaki10a})
and the second term is the joint posterior distribution of \textbf{$\boldsymbol{f}$
}and $\boldsymbol{f}^{'}$. The second term can be computed by
\begin{equation}
p(\boldsymbol{f},\boldsymbol{f}^{\boldsymbol{'}}\gv X,\boldsymbol{y},X_{s},\boldsymbol{s})=\frac{1}{Z}p(\boldsymbol{f},\boldsymbol{f}^{'}\gv X,X_{s})p(\boldsymbol{y}\gv\boldsymbol{f})p(\boldsymbol{s}\gv\boldsymbol{f^{'}})\label{eq:joint posterior}
\end{equation}
where $Z$ is a normalization term and $p(\boldsymbol{f},\boldsymbol{f}^{'}|X,X_{s})$
is the joint prior between \textbf{$\boldsymbol{f}$} and $\boldsymbol{f}^{'}$
which can be computed by
\begin{equation}
p(\boldsymbol{f},\boldsymbol{f}^{'}\gv X,X_{s})=\mathcal{N}\left(\boldsymbol{f}_{joint}\gv\boldsymbol{0},K_{joint}\right)\label{eq:joint prior}
\end{equation}
where $\boldsymbol{f}_{joint}=\left[\begin{array}{c}
\boldsymbol{f}\\
\boldsymbol{f}^{'}
\end{array}\right]$, $K_{joint}=\left[\begin{array}{cc}
K_{XX} & K_{XS}\\
K_{SX} & K_{SS}
\end{array}\right],$$K_{XX}$ and $K_{SS}$ are the self-covariance matrix of $X$ and
$X_{s}$, respectively and $K_{XS}$ is the covariance matrix between
$X$ and $X_{s}$. 

In Eq.(\ref{eq:joint posterior}), $p(\boldsymbol{s}|\boldsymbol{f^{'}})$
is the likelihood of derivative sign conditioning on derivative value.
Therefore, one has to build the link between derivative sign $\boldsymbol{s}$
and derivative value $\boldsymbol{f}^{'}$ in order to compute Eq.(\ref{eq:joint posterior})
. Riihimaki and Vehtari \cite{pmlr-v9-riihimaki10a} suggest using
a probit function to represent the likelihood of derivative signs
over latent derivative values as,
\begin{equation}
p(\boldsymbol{s}\gv\boldsymbol{f})=\prod_{i=1}^{m}\Phi\left(\frac{s_{i}\partial f^{(i)}}{\partial x_{d}^{(i)}}\frac{1}{\nu}\right)\label{eq:probit}
\end{equation}
where $\Phi(z)=\int_{-\infty}^{z}\mathcal{N}(x\gv0,1)dx$ and the
steepness $\nu$ indicates the consistency between the derivative
values and derivative signs. If we are confident about the derivative
signs, we set $\nu$ as a small value, otherwise large. Since the
likelihood in Eq.(\ref{eq:probit}) is not Gaussian, Eq.(\ref{eq:joint posterior})
is intractable analytically. Similar with the GP classification \cite{Rasmussen:2005:GPM:1162254},
Riihimaki and Vehtari \cite{pmlr-v9-riihimaki10a} used expectation
propagation (EP) \cite{Minka:2001:EPA} to approximate Eq.(\ref{eq:joint posterior}).
Briefly, we can use EP to approximate Eq.(\ref{eq:joint posterior})
as
\begin{align*}
 & q(\boldsymbol{f},\boldsymbol{f}^{'}\gv X,\boldsymbol{y},X_{s},\boldsymbol{s})\\
 & =\frac{1}{Z}p(\boldsymbol{f},\boldsymbol{f}^{'}\gv X,X_{s})p(\boldsymbol{y}\gv\boldsymbol{f})\prod_{i=1}^{N}t_{i}(f_{i}\gv Z_{i},\mu_{i},\sigma_{i})
\end{align*}
where $t_{i}(f_{i}\gv\tilde{Z}_{i},\tilde{\mu}_{i},\tilde{\sigma}_{i}^{2})\simeq\tilde{Z}_{i}\mathcal{N}(f_{i}\gv\tilde{\mu}_{i},\tilde{\sigma}_{i}^{2})$,
which defines a un-normalized Gaussian function with site parameter
$\tilde{Z}_{i}$, $\tilde{\mu_{i}}$ and $\tilde{\sigma}_{i}^{2}$.
Therefore Eq.(\ref{eq:joint posterior}) would be a product of multiple
Gaussian distributions after approximation. The detail inference can
be found in \cite{pmlr-v9-riihimaki10a}. Then the predictive mean
and variance of GP with derivative signs in Eq.(\ref{eq:predictive})
can be derived and they have the similar form with those in the standard
GP. 

If we set derivative signs with respect to one variable to be always
negative or positive, the resulted Gaussian process will be modeled
towards the desired monotonic shape on this variable. We denote it
as \emph{monotonic GP}. Usually the higher the number of sign observations
(that is a larger $m$), stronger is the monotonicity imposition.
However, due to the complexity $O((t+m)^{3})$ in GP with derivative
signs, it is not practical working with many derivative signs. In
our experiments, we place about five derivative signs per monotonic
dimension equally spaced within the bound of the variable.  

\subsection{Bayesian Optimization \label{subsec:Bayesian-Optimization}}

\begin{figure*}
\begin{centering}
\subfloat[]{\begin{centering}
\includegraphics[width=0.25\textwidth]{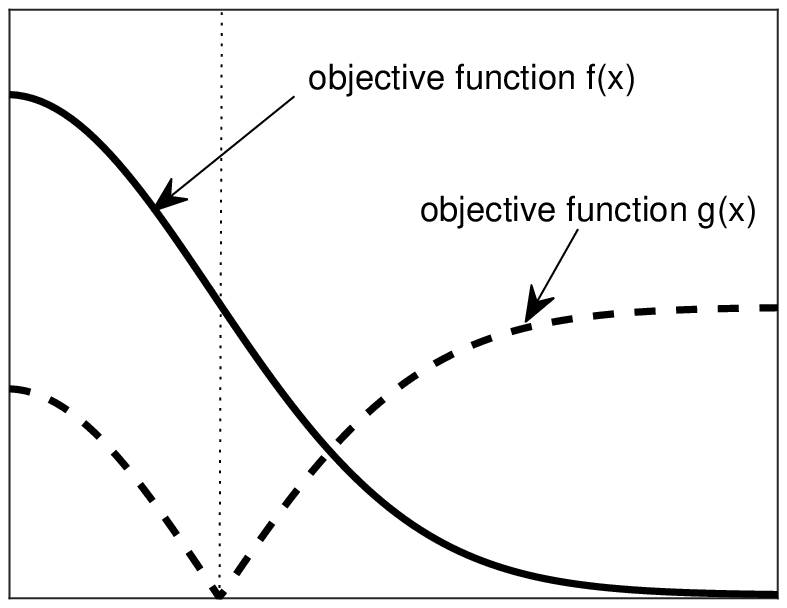}
\par\end{centering}
}\hspace{-0.5cm}\subfloat[]{\begin{centering}
\includegraphics[width=0.25\textwidth]{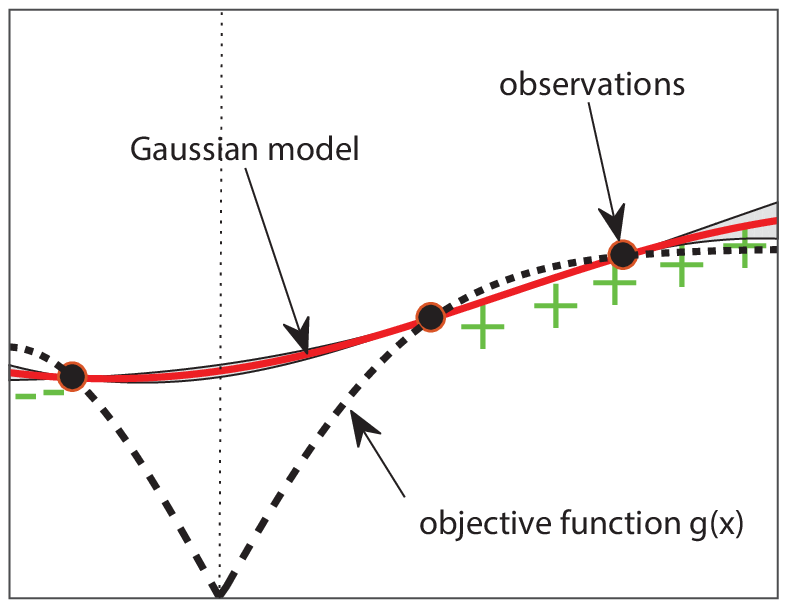}
\par\end{centering}
}\hspace{-0.5cm}\subfloat[]{\begin{centering}
\includegraphics[width=0.25\textwidth]{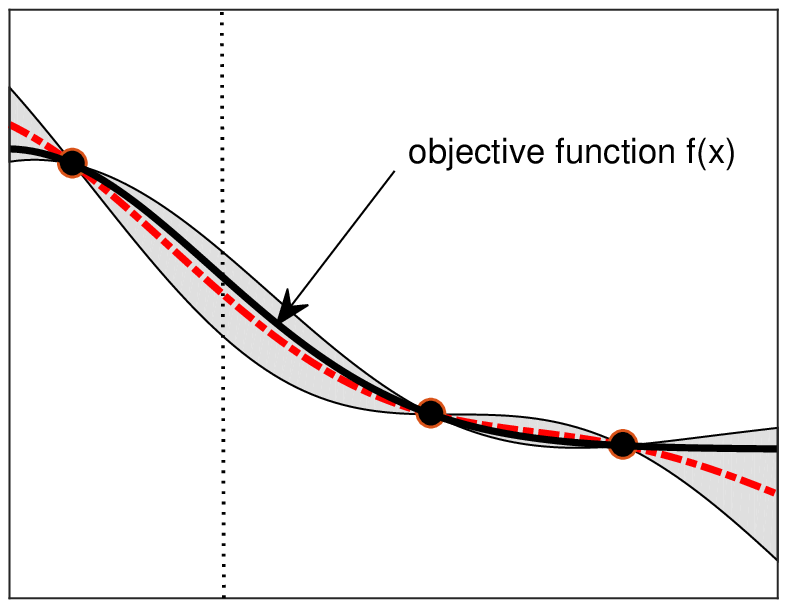} 
\par\end{centering}
}\hspace{-0.5cm}\subfloat[]{\begin{centering}
\includegraphics[width=0.25\textwidth]{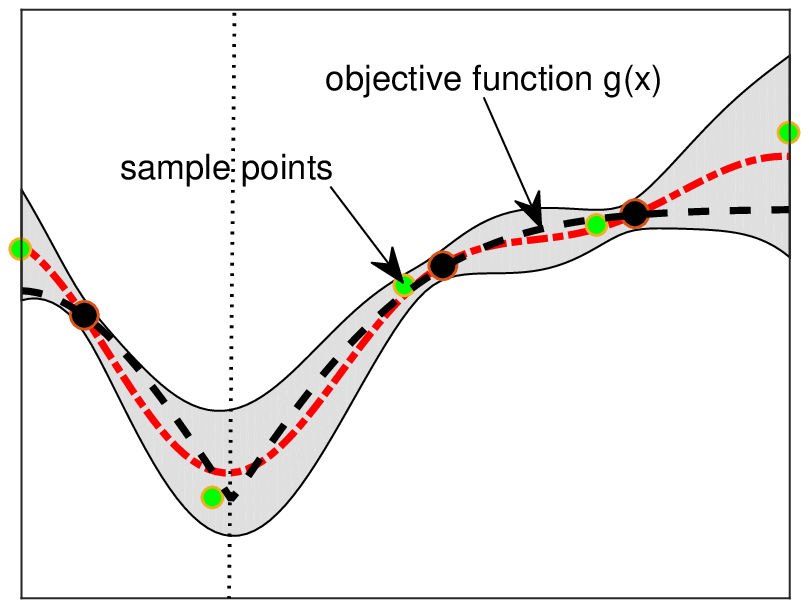}
\par\end{centering}
}
\par\end{centering}
\caption{\label{fig:posteriorGP}Illustration of the problem and solutions.
(a) Objective function $f(\boldsymbol{x})$ is monotonically decreasing
and $g(\boldsymbol{x})=|f(\boldsymbol{x})-y_{T}|$. The vertical dotted
line is the location of the target $y_{T}$. (b) \emph{BO-DS}: Posterior
GP of $g(\boldsymbol{x})$. The red dotted line represents the mean
function and the shadow represents predicted variance. The derivative
signs of $g(\boldsymbol{x})$ are derived based on the monotonicity
of $f(\boldsymbol{x})$. Information about derivative sign is lacking
in regions as discussed in text; (c) \emph{BO-MG:} Posterior GP of
$f(\boldsymbol{x})$, incorporating knowledge that $f(\boldsymbol{x})$
is monotonically decreasing. (d) \emph{BO-MG:} Posterior GP of $g(\boldsymbol{x})$
combining points sampled from GP in (c) and actual observations. }
\end{figure*}
Bayesian optimization (BO) is an efficient tool to globally optimize
an expensive black-box function. It is a greedy search procedure guided
by a surrogate function that is analytical and cheap to evaluate.
Typically, we use Gaussian process to model the latent function in
BO. The posterior mean and variance at each point can be analytically
derived based on Eq.(\ref{eq:predictive distribution-2}). Then the
surrogate function (or called acquisition function) is constructed
using both the predictive mean and variance. The next sample location
$\boldsymbol{x}_{t+1}$ is found by maximizing the acquisition function
and then $y_{t+1}$ is obtained after performing a new experiment
with $\boldsymbol{x}_{t+1}$. The new observation $\{\boldsymbol{x}_{t+1},y_{t+1}\}$
is augmented to update the GP. These steps are repeated till a satisfactory
outcome is reached or the iteration budget is exhausted. We present
a generic BO in Alg. \ref{alg:The-flow-ofBO}.

The acquisition function is designed to trade-off between exploitation
of high predictive mean and exploration of high epistemic uncertainty.
Choices of acquisition functions include Expected Improvement (EI)
\cite{srinivas10gaussian}, GP-UCB \cite{srinivas10gaussian}
and entropy search \cite{Hennig2012_entropy}. In this paper we use
GP-LCB for a minimization problem, which minimizes the acquisition
function
\begin{equation}
a_{t}(\boldsymbol{x})=\mu_{t-1}(\boldsymbol{x})-\sqrt{\alpha_{t}}\sigma_{t-1}(\boldsymbol{x})\label{eq:GP-LCB}
\end{equation}
where $\alpha_{t}$ is a positive trade-off parameter, $\mu_{t-1}(\boldsymbol{x})$
is the predicted mean and $\sigma_{t-1}(\boldsymbol{x})$ is the predicted
variance.

Simple regret at $t$th iteration is defined as $r_{t}=f(\boldsymbol{x}_{t})-f(\boldsymbol{x}^{*})$
for minimization problem where $\boldsymbol{x}^{*}$ is the global
optima of $f(\boldsymbol{x})$. Srinivas et al. \cite{srinivas10gaussian}
theoretically analyzed the regret bound of BO using the GP-LCB acquisition
function and showed that a) Bayesian optimization with GP-LCB is a
no-regret algorithm and b) the cumulative regret ($R_{T}=\sum_{t=1}^{T}r_{t}$)
grows only sub-linearly, \emph{i.e.} the convergence rate is the fastest
among all global optimizers known so far.\begin{algorithm}[t]
\caption{The standard Bayesian Optimization} \label{alg:The-flow-ofBO}  
\begin{algorithmic}[1]   
\FOR {$t=1,2\cdots$} 
\STATE Optimize for the next point $\boldsymbol{x}_{t+1}$$\leftarrow$$\text{argmax}_{\boldsymbol{x}_{t+1\in\boldsymbol{X}}}a(\boldsymbol{x}\gv\mathcal{D}_{1:t})$  
\STATE Evaluate the value $y_{t+1}$ 
\STATE Augment the data $\mathcal{D}_{1:t+1}=\{\mathcal{D}_{1:t},\{\boldsymbol{x}_{t+1},y_{t+1}\}\}$ 
\STATE Update the kernel matrix $\mathbf{K}$ 
\ENDFOR
\end{algorithmic}    \end{algorithm}

\section{Bayesian Optimization with Monotonicity Information\label{sec:framework}}

\begin{figure*}
\begin{centering}
\includegraphics[width=0.25\textwidth]{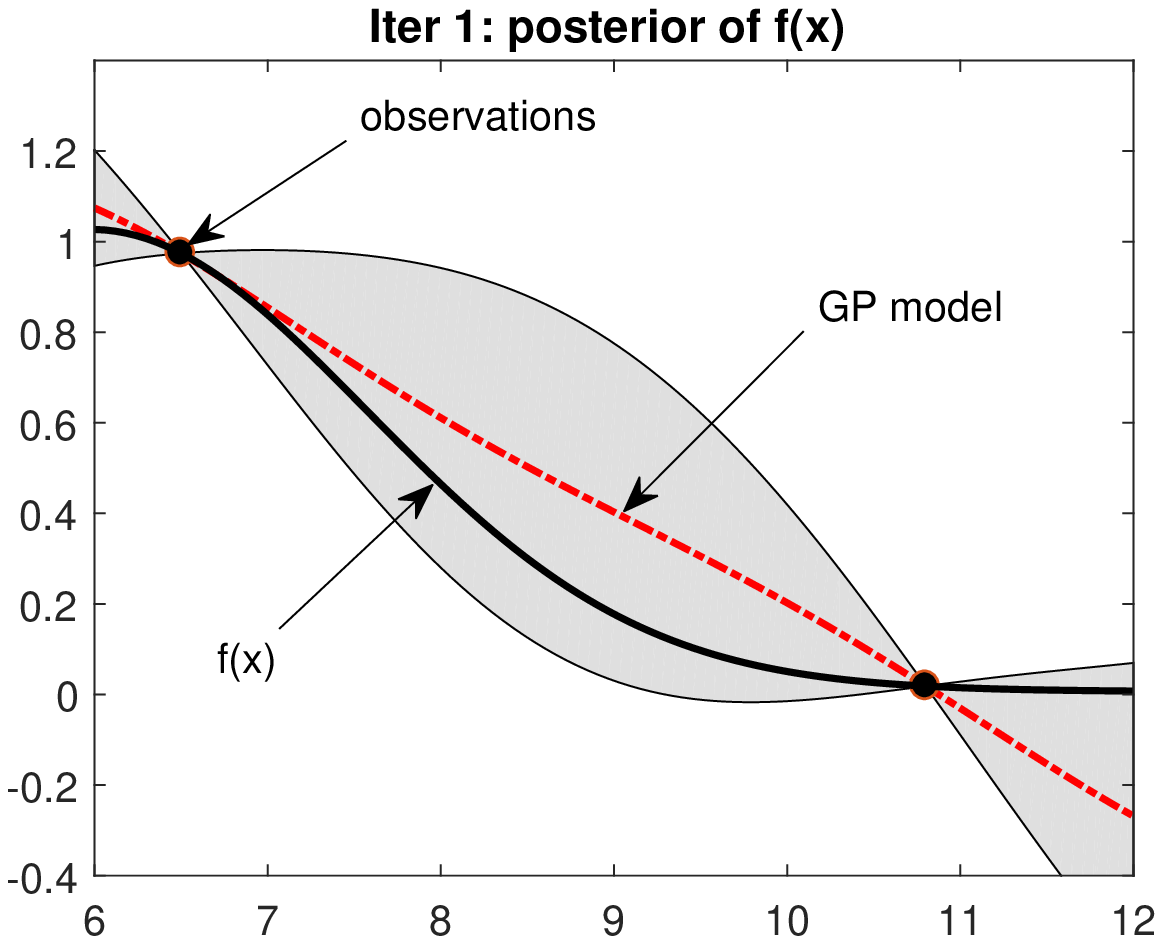}\includegraphics[width=0.25\textwidth]{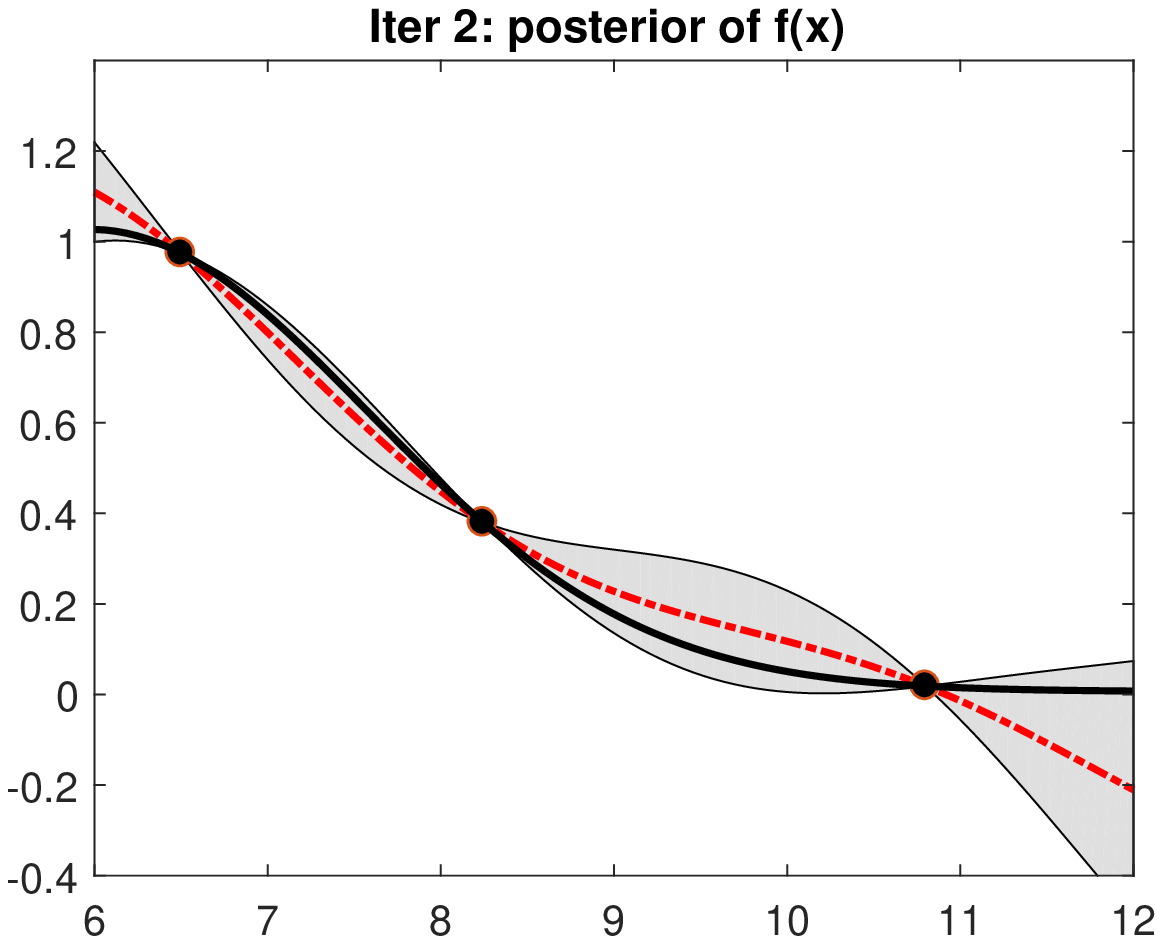}\includegraphics[width=0.25\textwidth]{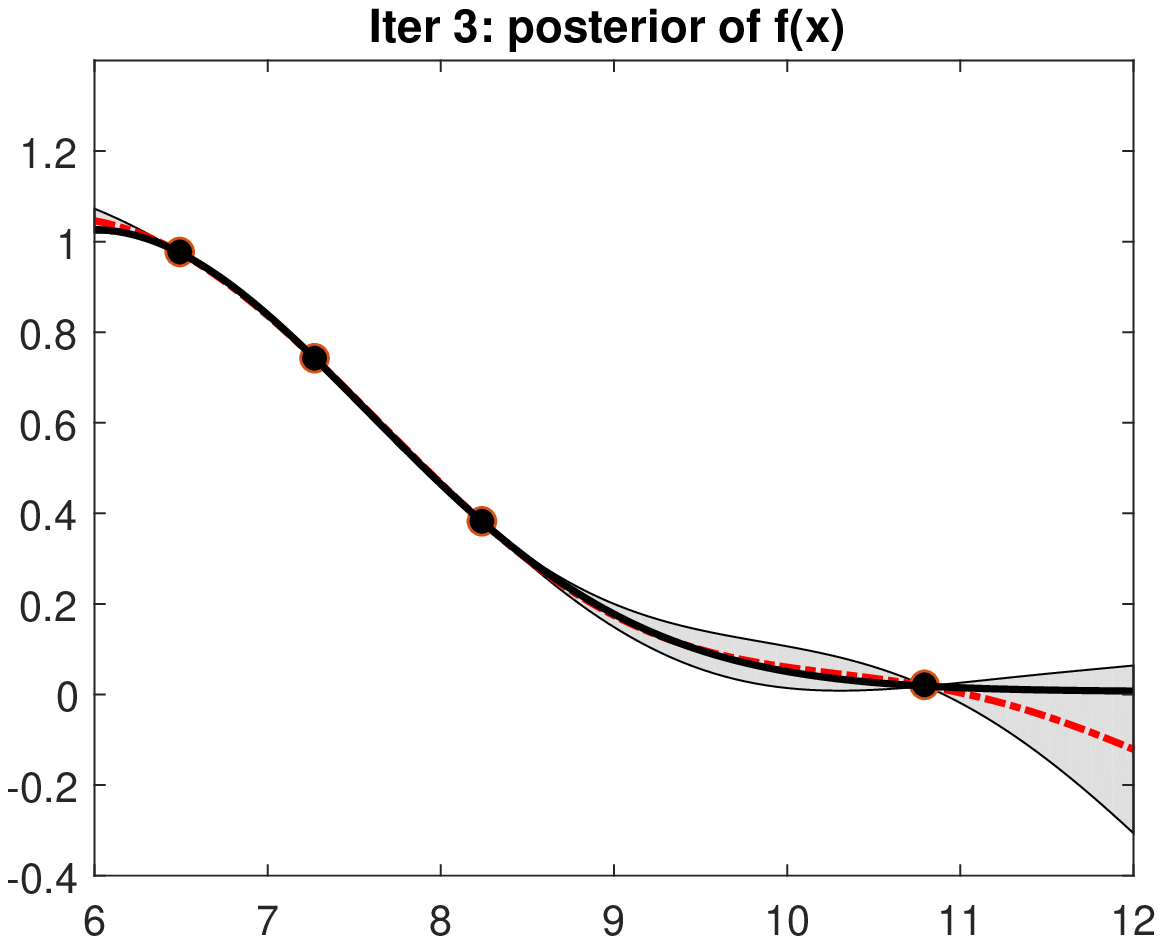}\includegraphics[width=0.25\textwidth]{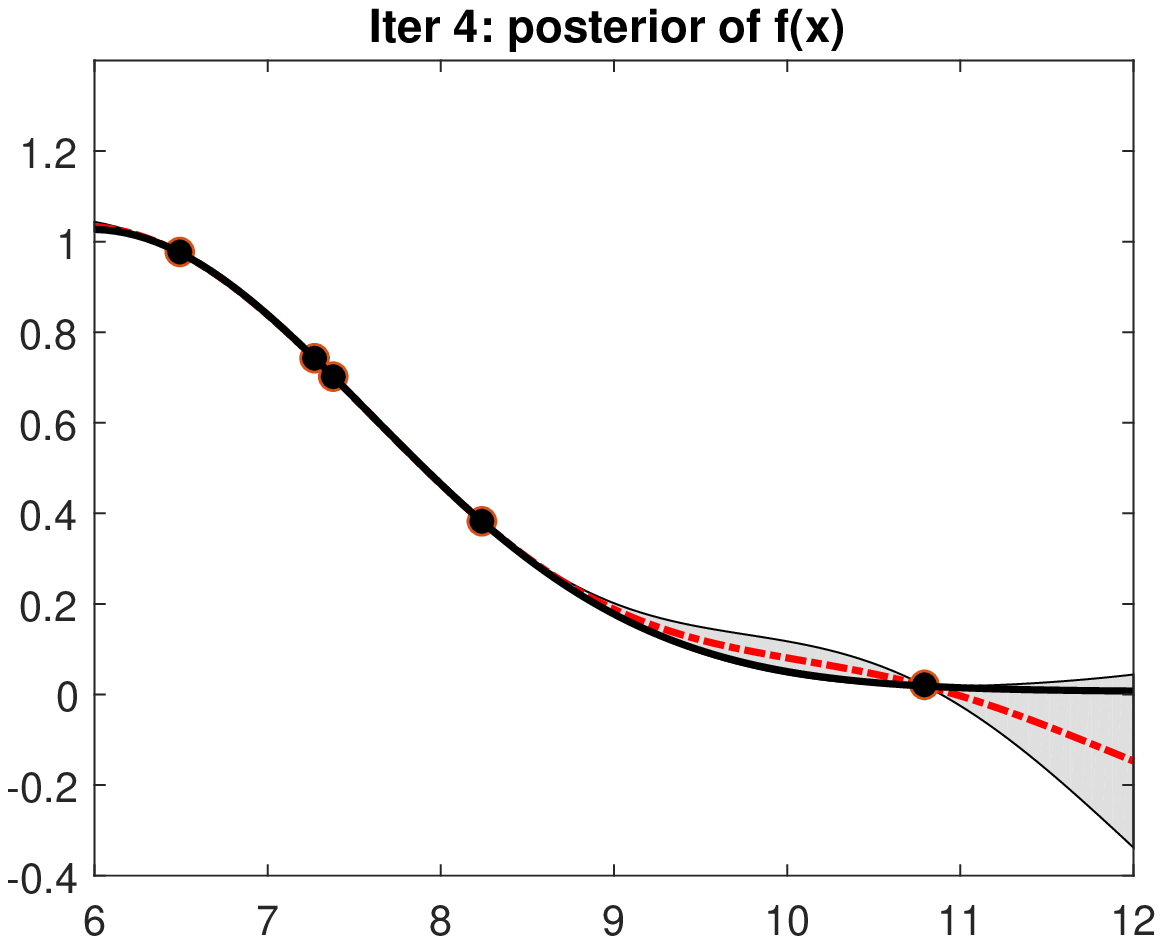}
\par\end{centering}
\begin{centering}
\includegraphics[width=0.25\textwidth]{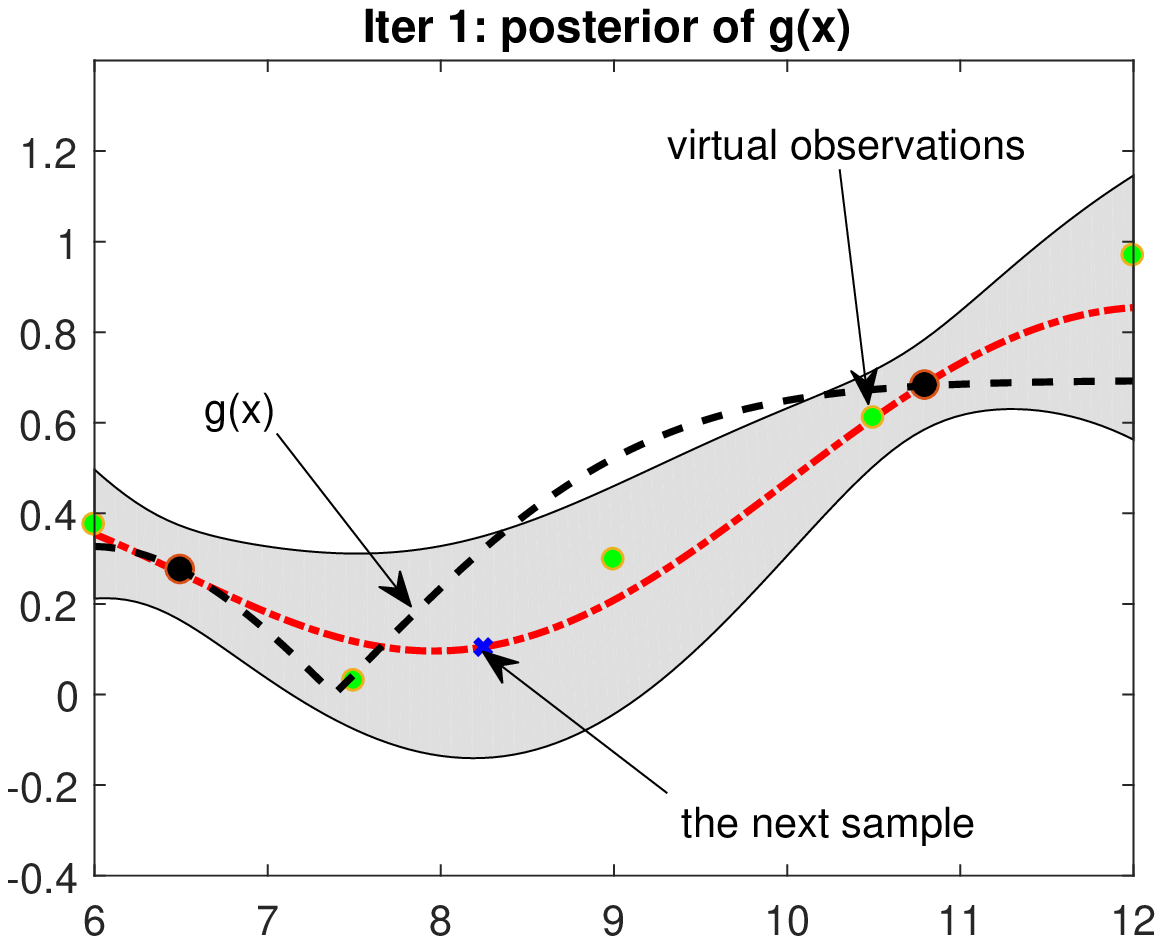}\includegraphics[width=0.25\textwidth]{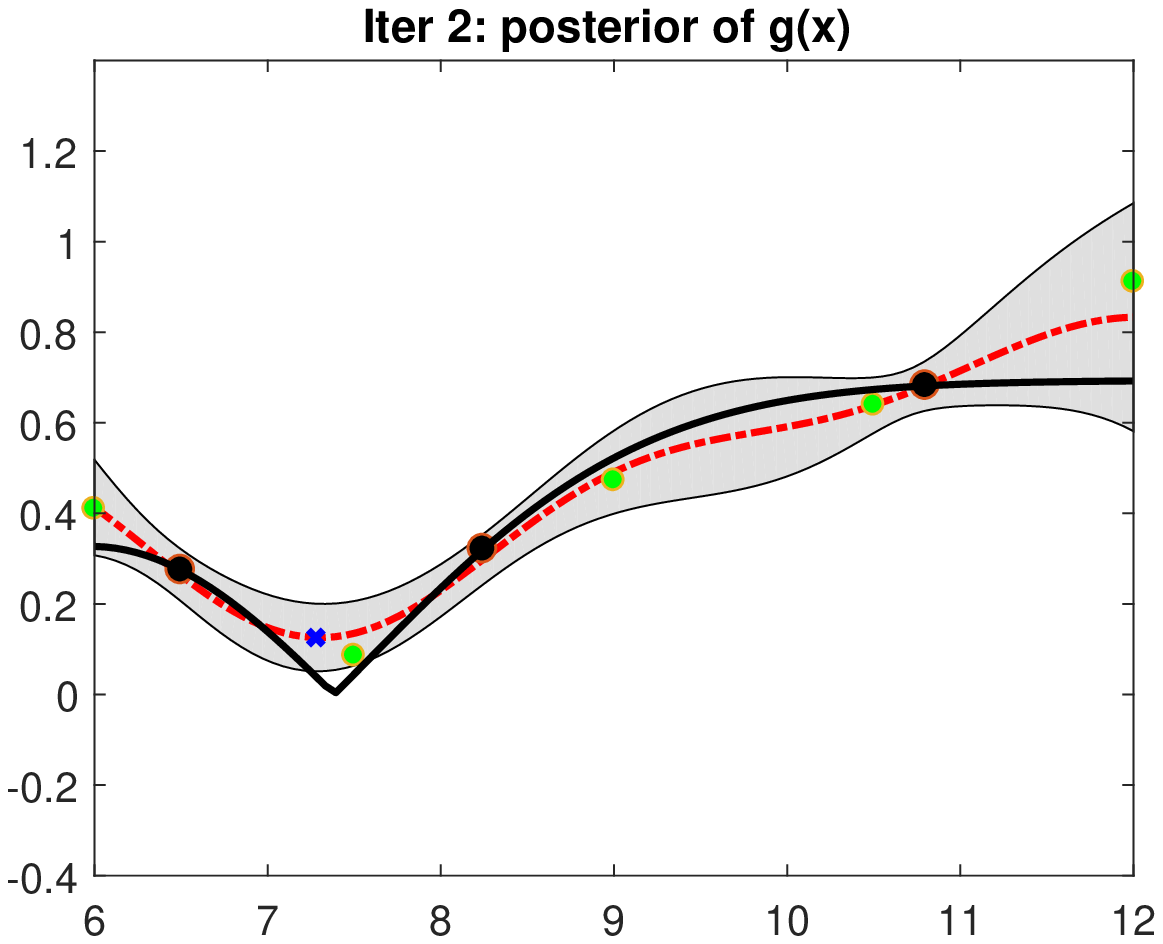}\includegraphics[width=0.25\textwidth]{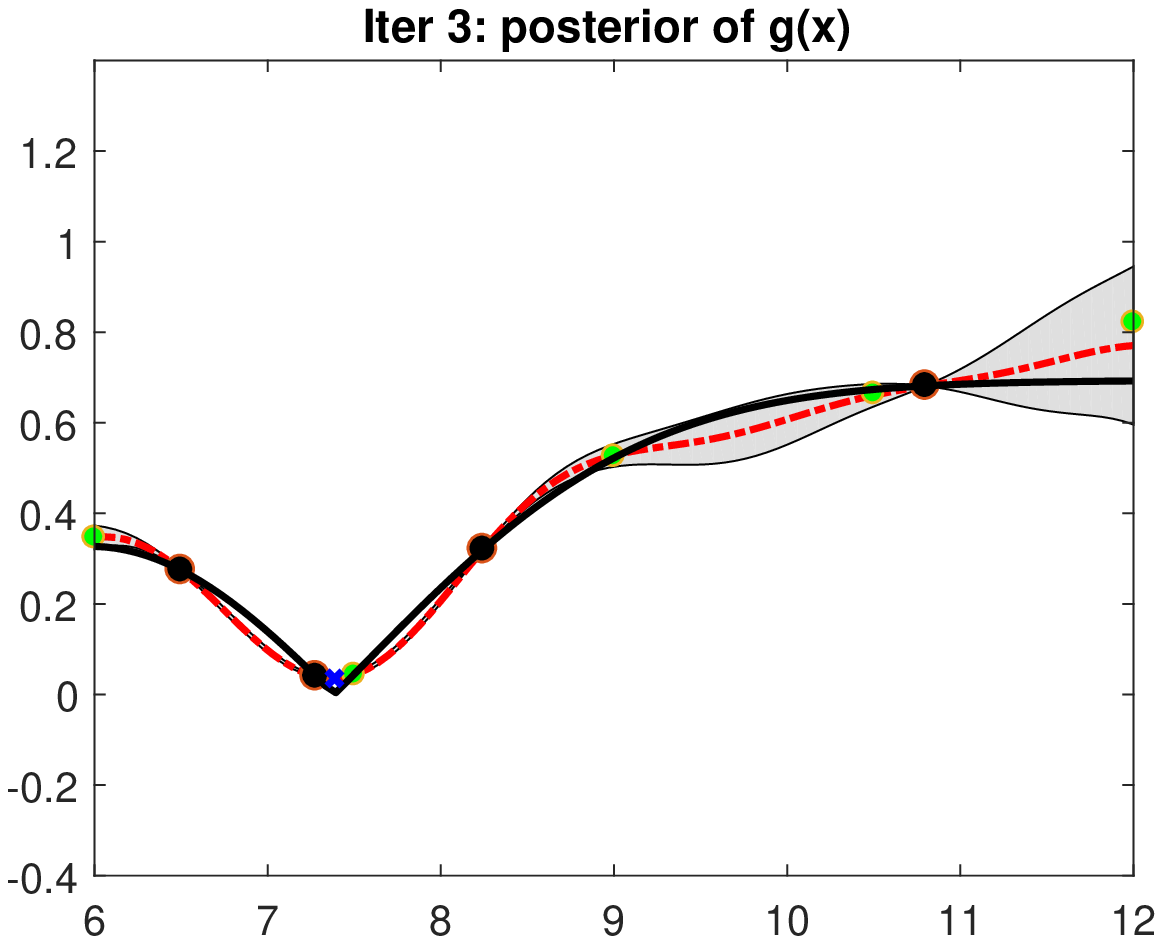}\includegraphics[width=0.25\textwidth]{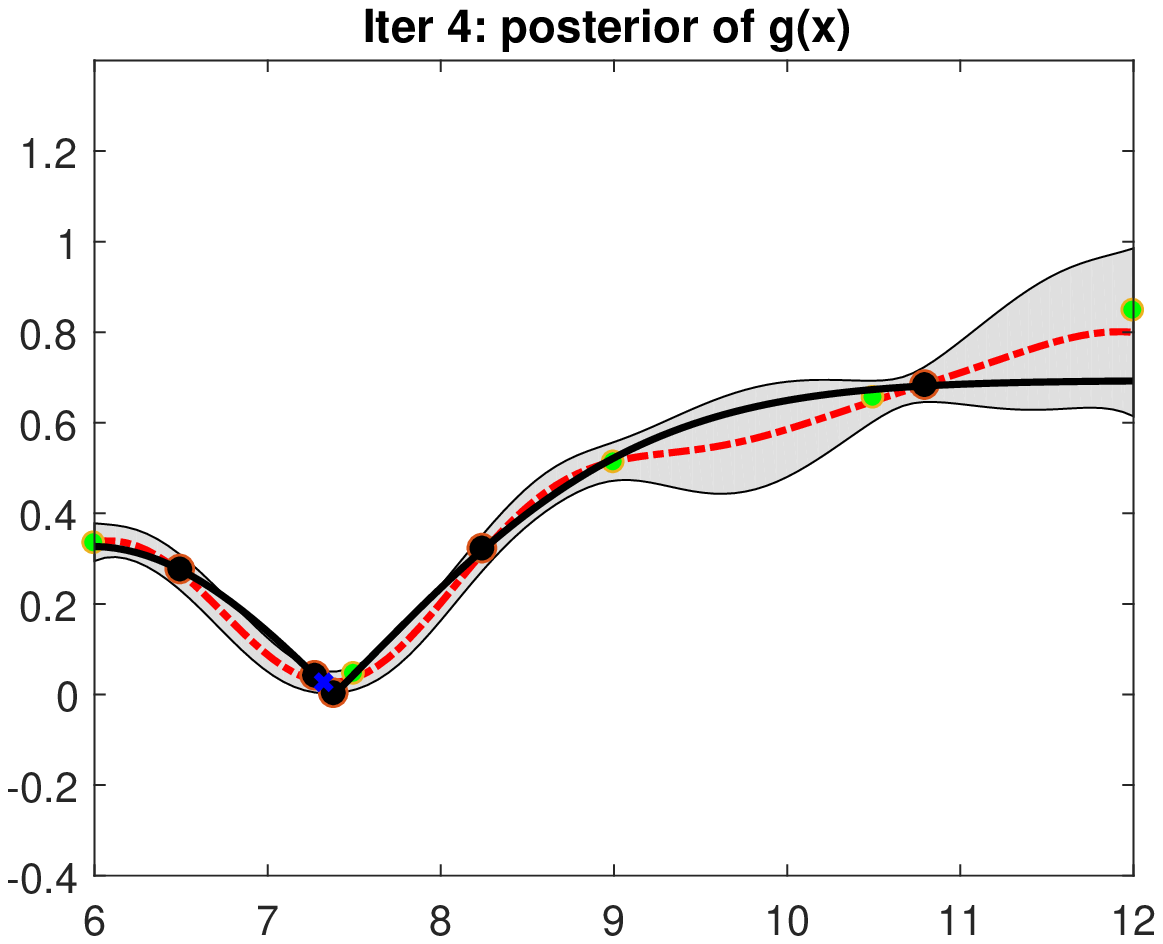}
\par\end{centering}
\caption{\label{fig:The-behavior-ofBO-MG}The behavior of BO-MG on a 1-d example
with $y_{T}=0.7$. The topmost plots shows the posterior of $f(\boldsymbol{x})$
by using monotonic GP. The plots in the bottom row depict the posterior
of $g(\boldsymbol{x})$ after introducing the virtual observations
sampled from the posterior GP of $f(\boldsymbol{x})$ (denoted by
green dots). The shade denotes the region covered by three times of
the predictive variance.}
\end{figure*}
Our objective is to reach a target value $y_{T}$ given the monotonicity
of $f(\boldsymbol{x})$. A natural choice is to minimize the difference
between the target and function values - Eq.(\ref{eq:objective function}).
We now discuss how to incorporate the monotonicity of $f(\boldsymbol{x})$
into BO to improve efficiency. \begin{algorithm}  
\caption{Bayesian optimization with derivative signs}
\label{alg:BO derivative signs} 
\begin{algorithmic}[1]  
\renewcommand{\algorithmicrequire}{\textbf{Input:}}  
\REQUIRE observations $\mathcal{D}_{1:t}=\{\boldsymbol{x_i},y_i\}_{i=1}^{t}$, the target value $y_{T}$, the monotonicity with respect to the $d$th variable.  

\FOR {$t=1,2,\cdots$}  
\STATE derive derivative sign observations $\mathcal{M}=\{\boldsymbol{x}_{s_i},s_{i}\}_{i=1}^{M}$ on $g(\boldsymbol{x})$  (Lemma \ref{lem:lemma1}).
\STATE obtain observations $\mathcal{G}=\{\boldsymbol{x}_i,|y_i-y_{T}|\}_{i=1}^{t}$;
\STATE build GP on $g(\boldsymbol{x})$ with $\mathcal{G}$ and $\mathcal{M}$ (Sec \ref{subsec:GP-Signs});
\STATE optimize for the next point $\boldsymbol{x}_{t+1}$$\leftarrow$$\text{argmax}_{\boldsymbol{x}_{t+1\in\mathcal{X}}}a(\boldsymbol{x}\gv\mathcal{G},\mathcal{M})$
\STATE evaluate the function $y_{t+1}=f(\boldsymbol{x}_{t+1})+\varepsilon$;
\STATE augment the data $\mathcal{D}_{1:t+1}=\{\mathcal{D}_{1:t},\{\boldsymbol{x}_{t+1},y_{t+1}\}\}$;
\ENDFOR

\end{algorithmic}   
\end{algorithm}

\subsection{\label{subsec:BO-DS}Bayesian Optimization with Derivative Signs
(BO-DS) }

A na\"ive method to utilize the monotonicity information is to derive
the property of $g(\boldsymbol{x})$ based on the given monotonicity
of $f(\boldsymbol{x})$ and locations of observations. We derive derivative
signs of $g(\boldsymbol{x})$ through the Lemma as follows:
\begin{lem}
\label{lem:lemma1}Let $f(\boldsymbol{x})$ be a monotonically decreasing
function with respect to the $d$th variable. Given the search bound
$[L_{d},U_{d}]$ of the $d$'th variable and an observation $\{\boldsymbol{x}_{i},y_{i}\}$($\boldsymbol{x}_{i}=[x_{i1},\cdots,x_{id},\cdots,x_{iD}]$),
if $y_{i}>y_{T}$, then $s<0$ at $\boldsymbol{x}_{s}=[x_{i1},\cdots,l_{s},\cdots,x_{iD}]$
for $\forall l_{s}\in[L_{d},x_{nd}]$ and if $y_{i}<y_{T}$, then
$s>0$ at $\boldsymbol{x}_{s}=[x_{i1},\cdots,l_{s},\cdots,x_{iD}]$
for $\forall l_{s}\in[x_{id},U_{d}]$.
\end{lem}
Proof: Since $f(\boldsymbol{x})$ is monotonically decreasing with
respect to the $d$th variable, then $f(\boldsymbol{x}_{s})>y_{i}$
for $l_{s}<x_{nd}$. Further we can get $|f(\boldsymbol{x}_{s})-y_{T}|>|y_{i}-y_{T}|$
if $y_{i}>y_{T}$. It means that $g(\boldsymbol{x}_{s})>g_{i}$ and
we can obtain the derivative sign $s<0$ at $\boldsymbol{x}_{s}$.
We can similarly prove the latter statement in Lemma 1. This lemma
is easy to extend to multiple dimensional case. 

Once a set of derivative signs $\mathcal{M}=\{\boldsymbol{x}_{s_{i}},s_{i}\}_{i=1}^{m}$
on $g(\boldsymbol{x})$ is acquired in this way, they are combined
with the actual observations $\mathcal{G}=\{\boldsymbol{x}_{i},|y_{i}-y_{T}|\}_{i=1}^{t}$.
Then a GP model can be constructed using the method of GP with derivative
signs in section \ref{subsec:GP-Signs} and BO is performed on $g(\boldsymbol{x})$
to acquire the next recommendation. We term this algorithm \emph{BO
with Derivative Signs (BO-DS)}, which is presented in Alg. \ref{alg:BO derivative signs}
.

A crucial drawback of this algorithm is that we do not know any derivative
information around the optimum, only away from it (See Figure \ref{fig:posteriorGP}(b)).
Thus we have only partially exploited the monotonicity information
of $f(\boldsymbol{x})$ in this approach. 

\subsection{\label{subsec:BO-MG}Bayesian Optimization with Monotonic GP (BO-MG)}

To overcome the drawback of BO-DS, we develop a two-stage algorithm
to eliminate the ambiguity of derivative signs in search bound. We
first model the mean function of posterior GP of $f(\boldsymbol{x})$
as a monotonic function,  then sample points from this GP and combine
them with existing actual observations to build a new GP model for
$g(\boldsymbol{x})$. Thus we make full use of the monotonicity of
$f(\boldsymbol{x})$ and transfer this critical knowledge to $g(\boldsymbol{x})$
through a set of sampled points. 

In detail, we model $f(\boldsymbol{x})$ using monotonic GP by placing
the consistent derivative signs $\{\boldsymbol{x}_{s},s\}$ across
the search space. We then sample $N$ points $X_{v}=\{\boldsymbol{x}_{j}^{v}\}_{j=1}^{N}$
from this monotonic GP. We denote the sampled set $\mathcal{V}=\{\boldsymbol{x}_{j}^{v},\mu_{f}(\boldsymbol{x}_{j}^{v}),\sigma_{f}^{2}(\boldsymbol{x}_{j}^{v})\}_{j=1}^{N}$
with the mean and variance. We note that it is important to retain
$\sigma_{f}^{2}(\boldsymbol{x}_{j}^{v})$ to maintain proper epistemic
uncertainty. Combining sampled points and existing observations $\mathcal{G}=\{\boldsymbol{x}_{i},|y_{i}-y_{T}|\}_{i=1}^{t}$,
we construct a new GP on $g(\boldsymbol{x})$ and then perform Bayesian
optimization. The mean and variance for a new point $\boldsymbol{x}_{t+1}$
in this GP are
\begin{align}
\mu_{g}(\boldsymbol{x}_{t+1})=\mathbf{k}^{T}K^{-1}[\boldsymbol{\mu}_{g}(X_{v});|\boldsymbol{y}-y_{T}|]\label{eq:monogp_mean}\\
\sigma_{g}^{2}(\boldsymbol{x}_{t+1})=k(\boldsymbol{x}_{t+1},\boldsymbol{x}_{t+1})-\mathbf{k}^{T}K^{-1}\mathbf{k}\label{eq:monogp_variance}
\end{align}
where $\boldsymbol{\mu}_{g}(X_{v})=|\boldsymbol{\mu}_{f}(X_{v})-y_{T}|$,
$\mathbf{k}=[k(\boldsymbol{x}_{t+1},\boldsymbol{x}_{1}^{v})\,\cdots\,k(\boldsymbol{x}_{t+1},\boldsymbol{x}_{N}^{v})\,k(\boldsymbol{x}_{t+1},\boldsymbol{x}_{1})\,\cdots\,k(\boldsymbol{x}_{t+1},\boldsymbol{x}_{t})]$
and
\begin{equation}
K=\left[\begin{array}{cc}
K_{VV} & K_{VX}\\
K_{XV} & K_{XX}
\end{array}\right]+\left[\begin{array}{cc}
\boldsymbol{\sigma}_{f}^{2}(X_{v}) & \boldsymbol{0}\\
\boldsymbol{0} & _{\sigma_{noise}^{2}}
\end{array}\right]\text{I}
\end{equation}
and $K_{VV}$ is the self-covariance matrix of $X_{v}$ and $K_{XV}$
is the covariance matrix between $X$ and $X_{v}$. The overall algorithm
is presented in Alg \ref{alg:BO monotonic GP}. The comparison between
BO-MG and BO-DS algorithms is illustrated in Figure \ref{fig:posteriorGP}.
To further show how BO-MG behaves we demonstrate this algorithm in
1-d example in Figure \ref{fig:The-behavior-ofBO-MG}. The BO-MG can
model the true mean function of $g(\boldsymbol{x})$ very well and
converge the optimum quickly.

A crucial step in BO-MG is to sample points from monotonic GP and
merge them with actual observations to build a new GP model, which
we denote as the combined GP. Adding sample points (virtual observations)
to the combined GP may reduce predictive variance. An undesirable
side effect is that it may result in the overconfidence in exploitation
due to the shrinkage of the epistemic uncertainty resulted from more
observations. To guarantee the algorithm's convergence, we need to
control for this overconfidence. If not corrected, it will avoid exploration
at the cost of exploitation for the combined GP. For the acquisition
function GP-LCB, a way to avoid overconfidence is to adjust the trade-off
parameter so that the exploration can be increased. We analyze the
setting of trade-off parameter in the next section. \begin{algorithm}  
\caption{Bayesian optimization with monotonic GP}
\label{alg:BO monotonic GP} 
\begin{algorithmic}[1]  
\renewcommand{\algorithmicrequire}{\textbf{Input:}}
\renewcommand{\algorithmiccomment}[1]{\hfill$\triangleright$\textit{#1}} 
\REQUIRE observations $\mathcal{D}_{1:t}=\{\boldsymbol{x_i},y_i\}_{i=1}^{t}$, the target value $y_{T}$, the monotonicity with respect to the $d$th variable
\FOR {$t=1,2,\cdots$}  
\STATE build monotonic GP on $f(\boldsymbol{x})$ using the consistent derivative signs  (Sec \ref{subsec:GP-Signs}); 
\STATE sample virtual observations $\mathcal{V}$  from the monotonic GP above (Sec \ref{subsec:BO-MG});
\STATE obtain observations $\mathcal{G}=\{\boldsymbol{x}_i,|y_i-y_{T}|\}_{i=1}^{t}$;
\STATE build GP on $g(\boldsymbol{x})$ using $\mathcal{V}$ and $\mathcal{G}$ (Sec \ref{subsec:BO-MG});
\STATE sample $\boldsymbol{x}_{t+1}$$\leftarrow$$\text{argmax}_{\boldsymbol{x}_{t+1\in\mathcal{X}}}a(\boldsymbol{x}\gv\mathcal{G},\mathcal{V})$;
\STATE evaluate the function $y_{t+1}=f(\boldsymbol{x}_{t+1})+\varepsilon$;
\STATE augment the data $\mathcal{D}_{1:t+1}=\{\mathcal{D}_{1:t},\{\boldsymbol{x}_{t+1},y_{t+1}\}\}$;
\ENDFOR

\end{algorithmic}   
\end{algorithm}

\subsection{\label{subsec:Theoretical-Analysis-for}Theoretical Analysis for
BO-MG}

We denote $g$ as a sample from the combined GP model. With $N_{1}$
sampled points, the GP-LCB decision rule for the next point is given
as
\begin{equation}
\boldsymbol{x}_{t}^{N_{1}}=\argmin{\boldsymbol{x}\in\mathcal{X}}\mu_{t}^{N_{1}}(\boldsymbol{x})-\sqrt{\alpha_{t}}\sigma_{t-1}^{N_{1}}(\boldsymbol{x})
\end{equation}
where $\mu_{t-1}^{N_{1}}(\boldsymbol{x})$ and $\sigma_{t-1}^{N_{1}}(\boldsymbol{x})$
are the predictive mean and variance in this GP. With $N_{2}$ ($N_{2}>N_{1}$
and $\boldsymbol{x}_{1:N_{1}}\subset\boldsymbol{x}_{1:N_{2}}$) sampled
points, the decision rule is
\begin{equation}
\boldsymbol{x}_{t}^{N_{2}}=\argmin{\boldsymbol{x}\in\mathcal{X}}\mu_{t-1}^{N_{2}}(\boldsymbol{x})-\sqrt{\beta_{t}}\sigma_{t-1}^{N_{2}}(\boldsymbol{x})
\end{equation}
where $\mu_{t-1}^{N_{2}}(\boldsymbol{x})$ and $\sigma_{t-1}^{N_{2}}(\boldsymbol{x})$
are corresponding predictive mean and variance. 

Suppose these two GPs use the same hyperparameters, then $\mu_{t-1}^{N_{1}}(\boldsymbol{x})$
is approximately equal to $\mu_{t-1}^{N_{2}}(\boldsymbol{x})$ and
$\sigma_{t-1}^{N_{2}}(\boldsymbol{x})$ is less than $\sigma_{t-1}^{N_{1}}(\boldsymbol{x})$
due to the introduction of sampled points for $\forall t$ and $\forall\boldsymbol{x}\in\mathcal{X}$.
To overcome the overconfidence in exploitation of the combined GP,
we must choose a proper $\beta_{t}$ to increase its confidence intervals
so that $\sqrt{\beta_{t}}\sigma_{t-1}^{N_{2}}(\boldsymbol{x})$ can
contain $\sqrt{\alpha_{t}}\sigma_{t-1}^{N_{1}}(\boldsymbol{x})$ for
$\forall t$ and $\forall\boldsymbol{x}\in\mathcal{X}$, i.e.
\begin{equation}
\sqrt{\beta_{t}}\sigma_{t-1}^{N_{2}}(\boldsymbol{x})\geq\sqrt{\alpha_{t}}\sigma_{t-1}^{N_{1}}(\boldsymbol{x})\label{eq:betaalpha}
\end{equation}
 We use the choice of $\alpha_{t}$ derived by \cite{srinivas10gaussian}.
The core task becomes to bound the ratio 
\begin{equation}
r_{t-1}(\boldsymbol{x})=\sigma_{t-1}^{N_{1}}(\boldsymbol{x})/\sigma_{t-1}^{N_{2}}(\boldsymbol{x})\label{eq:varianceratio}
\end{equation}
 As in \cite{desautels14parallelizing}, this ratio can be computed
by the proposition as follows:
\begin{prop}
The ratio of the standard deviation of the posterior over $g(\boldsymbol{x})$,
conditioned on observations $y_{1:t-1}$ and $N_{1}$ sampled points
to that when $g(\boldsymbol{x})$ is conditioned on observations $y_{1:t-1}$
and $N_{2}$ sampled points is
\begin{equation}
\frac{\sigma_{t-1}^{N_{1}}(\boldsymbol{x})}{\sigma_{t-1}^{N_{2}}(\boldsymbol{x})}=\exp\left(I(g(\boldsymbol{x});y_{(N_{1}+1):N_{2}}\gv y_{1:t-1}\cup y_{1:N_{1}}\right)
\end{equation}
\end{prop}
We prove it by expanding the mutual information as follows:
\begin{align*}
 & I(g(\boldsymbol{x});y_{(N_{1}+1):N_{2}}\gv y_{1:t-1}\cup y_{1:N_{1}})\\
 & =H(g(\boldsymbol{x})\gv y_{1:t-1}\cup y_{1:N_{1}})-H(g(\boldsymbol{x})\gv y_{1:t-1}\cup y_{1:N_{2}})\\
 & =\frac{1}{2}\log\left(2\pi e\sigma_{t-1}^{N_{1}}(\boldsymbol{x})\right)-\frac{1}{2}\log\left(2\pi e\sigma_{t-1}^{N_{2}}(\boldsymbol{x})\right)\\
 & =\log\left(\sigma_{t-1}^{N_{1}}(\boldsymbol{x})/\sigma_{t-1}^{N_{2}}(\boldsymbol{x})\right)
\end{align*}
It shows that there exists a constant $C$ such that $I(g(\boldsymbol{x});y_{(N_{1}+1):N_{2}}\gv y_{1:t-1}\cup y_{1:N_{1}})\leq C$
for $\forall t$ and $\forall\boldsymbol{x}\in\mathcal{X}$. Therefore
we can successfully bound $r_{t-1}(\boldsymbol{x})\leq\exp(C)$. 

By the monotonicity and submodularity properties of mutual information
\cite{desautels14parallelizing,Krause:2005:NNV:3020336.3020377},
we get:
\begin{align}
 & I(g(\boldsymbol{x});y_{(N_{1}+1):N_{2}}\gv y_{1:t-1}\cup y_{1:N_{1}})\nonumber \\
 & \leq I(g;y_{(N_{1}+1):N_{2}}\gv y_{1:t-1}\cup y_{1:N_{1}})\\
 & \leq\max_{\mathcal{A}\subseteq\mathcal{X},|\mathcal{A}|\leq N_{2}-N_{1}}I(g;y_{\mathcal{A}}\gv y_{1:t-1}\cup y_{1:N_{1}})\\
 & \leq\max_{\mathcal{A}\subseteq\mathcal{X},|\mathcal{A}|\leq N_{2}-N_{1}}I(g;y_{\mathcal{A}})=\gamma_{N_{2}-N_{1}}\label{eq:information gain}
\end{align}
Generally $\gamma_{N_{2}-N_{1}}$ is difficult to calculate since
it generally requires to compute the information gain for all combinations
of $(N_{2}-N_{1})$ points. Fortunately, Andreas and Carlos \cite{Krause:2005:NNV:3020336.3020377}
demonstrated an easy method to obtain upper bound on $\gamma_{N_{2}-N_{1}}$.
Specifically, they show 
\begin{equation}
\gamma_{N_{2}-N_{1}}\leq\frac{e}{e-1}I(g;y_{N_{2}-N_{1}})\label{eq:uncertainty sampling}
\end{equation}
where $I(g;y_{N_{2}-N_{1}})$ the information gain by observing the
set of observations $y_{(N_{1}+1):N_{2}}$ of the actions $\{\boldsymbol{x}_{N_{1}+1},\cdots,\boldsymbol{x}_{N_{2}}\}$
selected using uncertainty sampling \cite{desautels14parallelizing}.
It implies that we can use uncertainty sampling to select $N_{2}-N_{1}$
sampled points in BO-MG such that we can obtain $C$. With $C=\gamma_{N_{2}-N_{1}}$,
we can get the regret bound as follows: 
\begin{thm}
Let $\delta\in(0,1)$ and run BO-MG with GP-LCB decision rule with
$\beta_{t}=\exp(2C)\alpha_{t}$, we get a cumulative regret bound
$R_{T}$ with a high probability
\begin{equation}
\text{Pr}\{R_{T}\leq\sqrt{C_{1}T\exp(2\gamma_{N_{2}-N_{1}})\alpha_{t}\gamma_{T}}+2,\forall T\geq1\}=1-\delta\label{eq:regret bound}
\end{equation}
where $C_{1}=8/\log(1+\sigma_{noise}^{2})$, $\gamma_{T}$ is the
maximum information gain between the function values $f_{1:T}$ and
the noisy observations $y_{1:T}$ , $\gamma_{N_{2}-N_{1}}$ is defined
in Eq.(\ref{eq:information gain}), and $\alpha_{t}=2\log(2t^{2}\pi^{2}/(3\delta))+2d\log\left(dt^{2}bl\sqrt{\log(4da/\delta)}\right)$. 
\end{thm}
The proof is similar to that in \cite{srinivas10gaussian}.

\paragraph*{Discussion}

We have explicitly discussed that the convergence rate of \emph{BO-MG}
can be guaranteed if $\beta_{t}=\exp(2C)\alpha_{t}$ and $C=\gamma_{N_{2}-N_{1}}$.
In practice, $N_{1}$ can be a very small one and then the maximum
information gain $\gamma_{N_{2}-N_{1}}$ grows with the size of $N_{2}$
and $C$ would be very large and thus the algorithm tends to over-explore
if we use the computed $C$ for $\beta_{t}$. Fortunately we can also
set $\beta_{t}=\left(\max\left(r_{t-1}(\boldsymbol{x})\right)\right)^{2}\alpha_{t}$
in order to guarantee $\beta_{t}\geq r_{t-1}^{2}(\boldsymbol{x})\alpha_{t}$
(Eq. \ref{eq:betaalpha}) for $\forall t$ and $\forall\boldsymbol{x}\in\mathcal{X}$.
Actually we can obtain the maximal value of $r_{t-1}(\boldsymbol{x})$
by maximizing Eq.(\ref{eq:varianceratio}) for $\forall\boldsymbol{x}\in\mathcal{X}$
at iteration $t$. In this way we can guarantee the convergence of
BO-MG. For good practical performance, a more aggressive method is
to reduce $\beta_{t}$ by a correction factor $\eta$ \cite{srinivas10gaussian}
\begin{equation}
\beta_{t}=\left(\max\left(r_{t-1}(\boldsymbol{x})\right)\right)^{2}\eta\alpha_{t}\label{eq:tradeoff-MG}
\end{equation}
Eq.(\ref{eq:varianceratio}) indicates that the value $\max\left(r_{t-1}(\boldsymbol{x})\right)$
is increasing with $N_{2}$ (assume $N_{1}$ is fixed) and therefore
we can adjust $\eta$ for different $N_{2}$ for better practical
performance.

\section{Experiments\label{sec:experiments}}

We compare our proposed method with the following algorithms: 
\begin{itemize}
\item Bayesian optimization with monotonic GP (\emph{BO-MG}) which incorporates
the sampled points from the monotonic GP into Bayesian optimization
(Alg. \ref{alg:BO monotonic GP});
\item Bayesian optimization with derivative signs (\emph{BO-DS}) which directly
incorporates the derivative signs derived from prior monotonicity
into BO (Alg. \ref{alg:BO derivative signs});
\item standard Bayesian optimization (\emph{standard BO}) which does not
include any prior knowledge (Alg. \ref{alg:The-flow-ofBO}).
\end{itemize}
For all three algorithms, we automatically estimate the hyperparameters
of the SE kernel in GP including the length scale $l$ and the output
variance $\epsilon$ and the noise variance $\sigma_{noise}^{2}$
at each iteration. Both \emph{BO-DS} and \emph{BO-MG} requires the
GP with derivative signs. We empirically set $\nu=0.01$ and used
the GPstuff toolbox \cite{Vanhatalo:2013:GBM:2567709.2502617} to
implement the GP with derivative signs. The acquisition function we
used for all algorithms is the GP-LCB. For standard BO and BO-DS,
the trade-off parameter $\alpha_{t}$ in Eq.(\ref{eq:GP-LCB}) can
be set by following \cite{srinivas10gaussian} but is scaled down
with a small factor as \cite{srinivas10gaussian} and \cite{desautels14parallelizing}
did (we use 0.1 in our experiments). For BO-MG, we used the trade-off
parameter $\beta_{t}$ in Eq.(\ref{eq:tradeoff-MG}). To compute $\max\left(r_{t-1}(\boldsymbol{x})\right)$
we sampled $N_{1}=5$, $N_{2}=10$ for 2D functions, $N_{1}=5$, $N_{2}=20$
for 5D functions and $N_{1}=5$, $N_{2}=40$ for 7D functions using
Latin hypercube sampling and ensured sampled points $\boldsymbol{x}_{1:N_{1}}\subset\boldsymbol{x}_{1:N_{2}}$.
BO-MG provides competitive performance with $\eta=0.1$ for 1D\textasciitilde 5D
functions and $\eta=0.01$ for 7D functions in our experiments. We
run experiments for 20 trials with random initial points and report
the average mean and the standard error. The code is available in
https://bit.ly/2sDFQ35.

We first compared algorithms on the optimization of benchmark functions
and hyperparameter tuning in neural network. We then solved two real-world
applications - the optimization of short fibers with targeted length
and porous architecture (scaffold) design for biomaterials with target
porosity using 3D printing. 

\subsection{Optimization of benchmark functions}

We optimize the following benchmark functions: 
\begin{figure*}
\begin{centering}
\subfloat[2D function $f_{1}$. ]{\centering{}\includegraphics[width=0.33\textwidth]{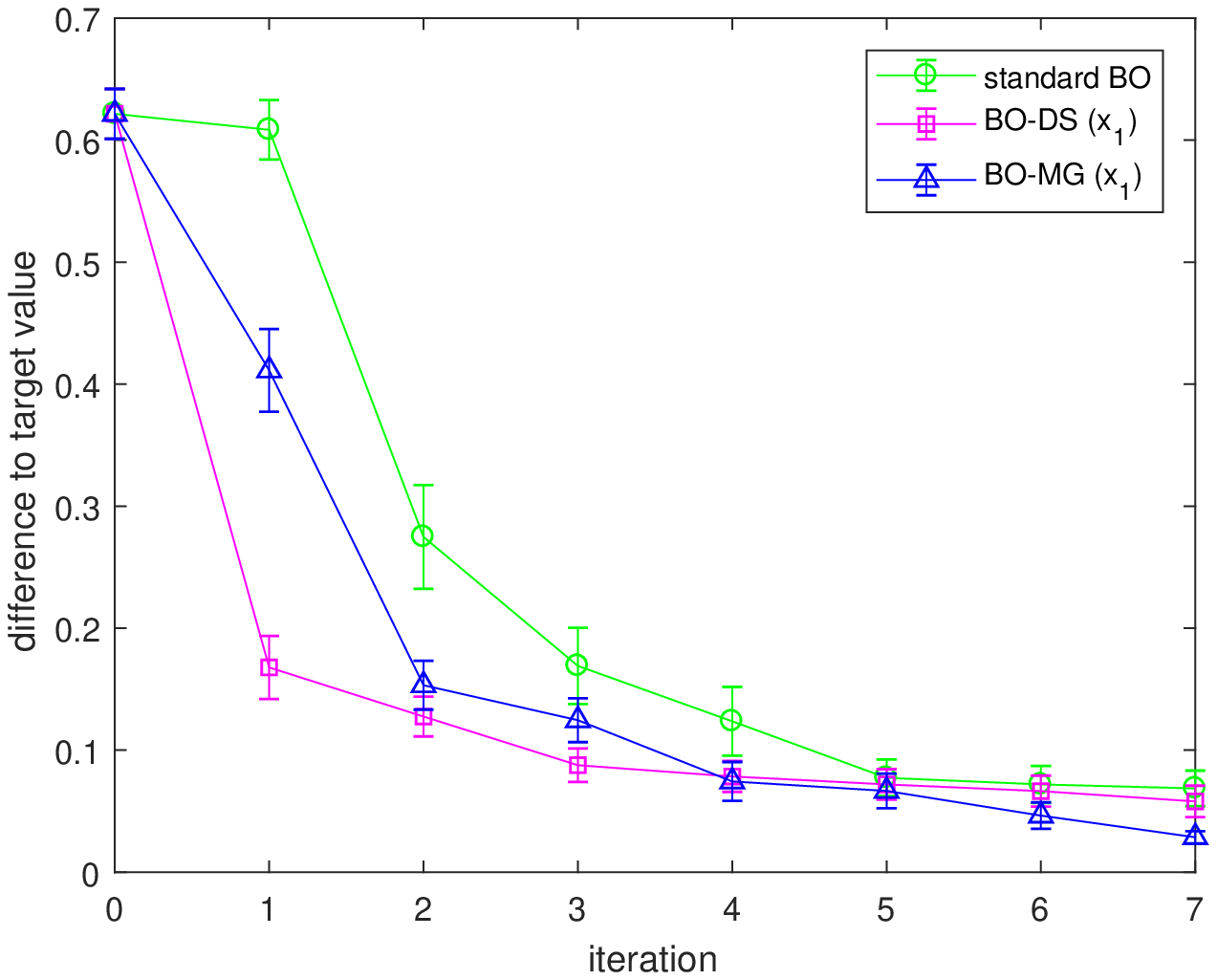}}\subfloat[5D function $f_{2}$.]{\centering{}\includegraphics[width=0.33\textwidth]{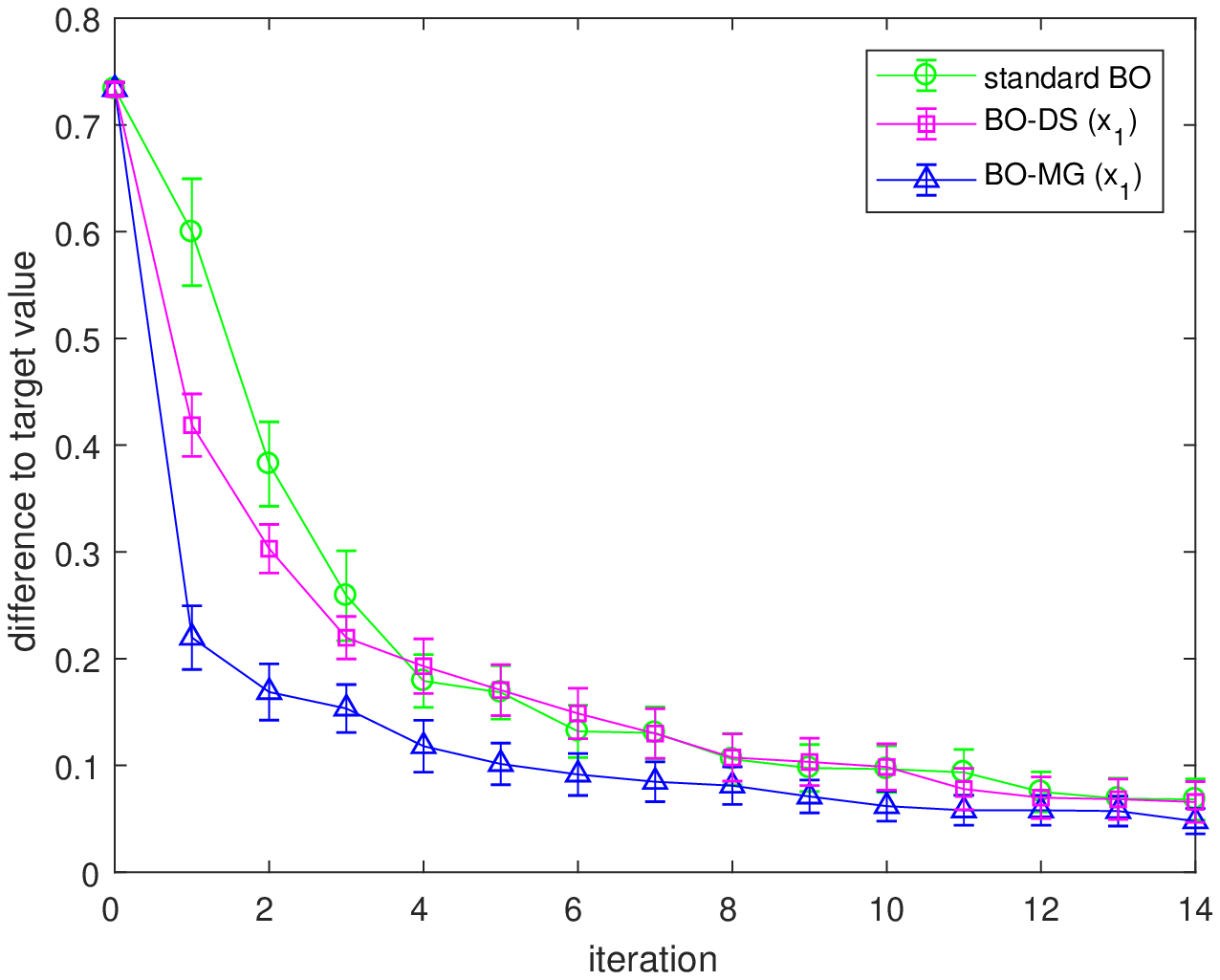}}\subfloat[7D function $f_{3}$.]{\centering{}\includegraphics[width=0.33\textwidth]{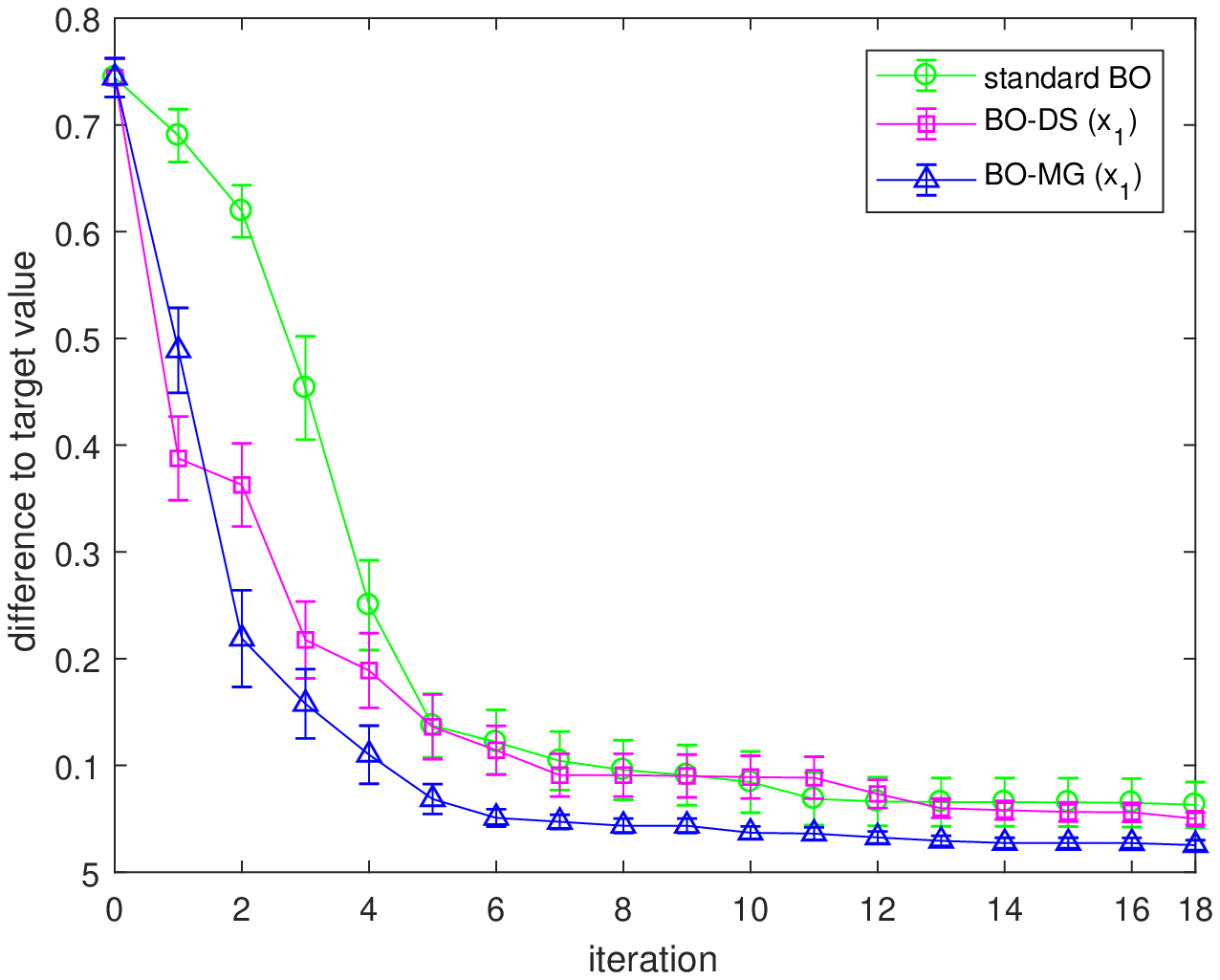}}
\par\end{centering}
\begin{centering}
\subfloat[2D function $f_{4}$. ]{\centering{}\includegraphics[width=0.33\textwidth]{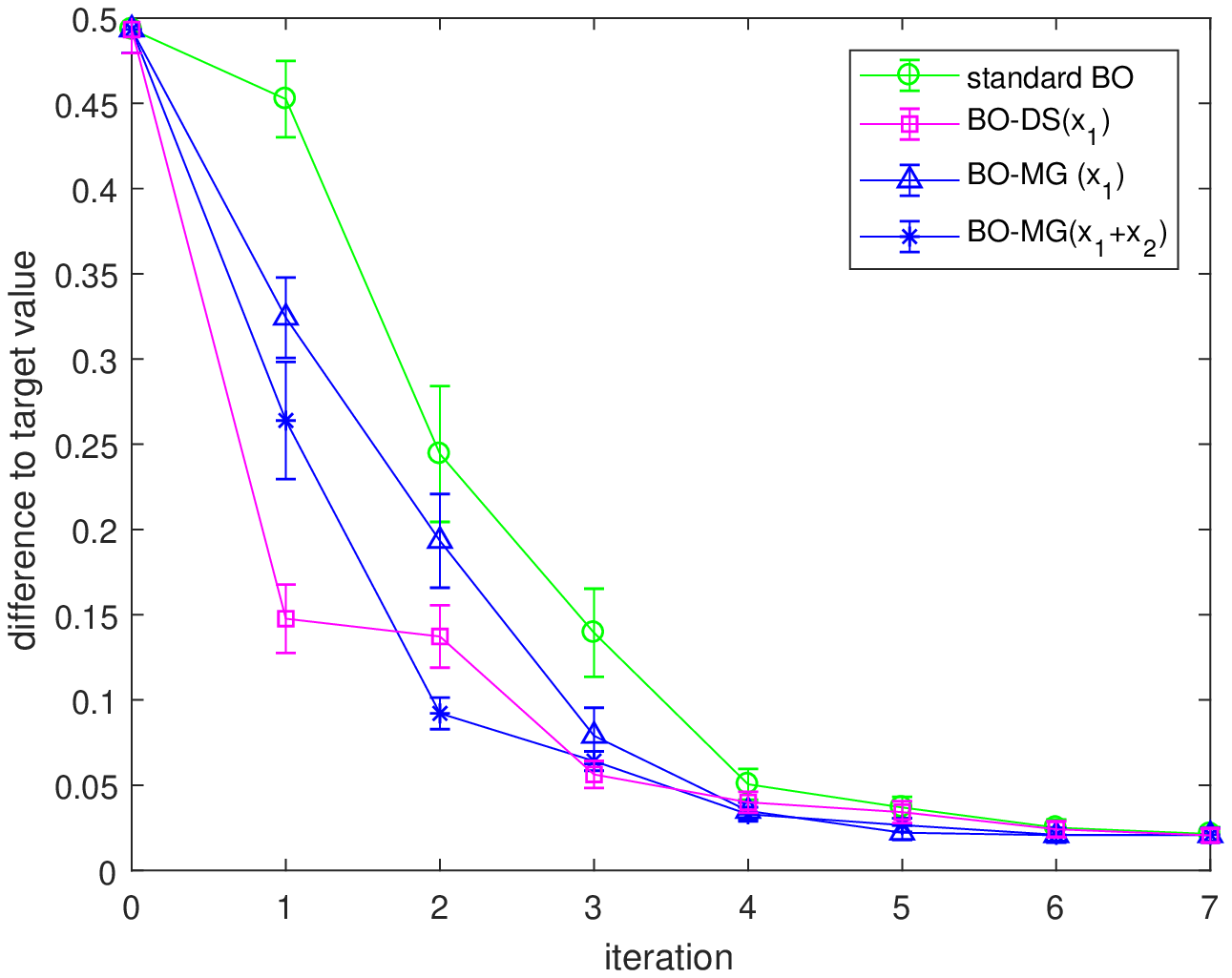}}\subfloat[5D function $f_{5}$.]{\centering{}\includegraphics[width=0.33\textwidth]{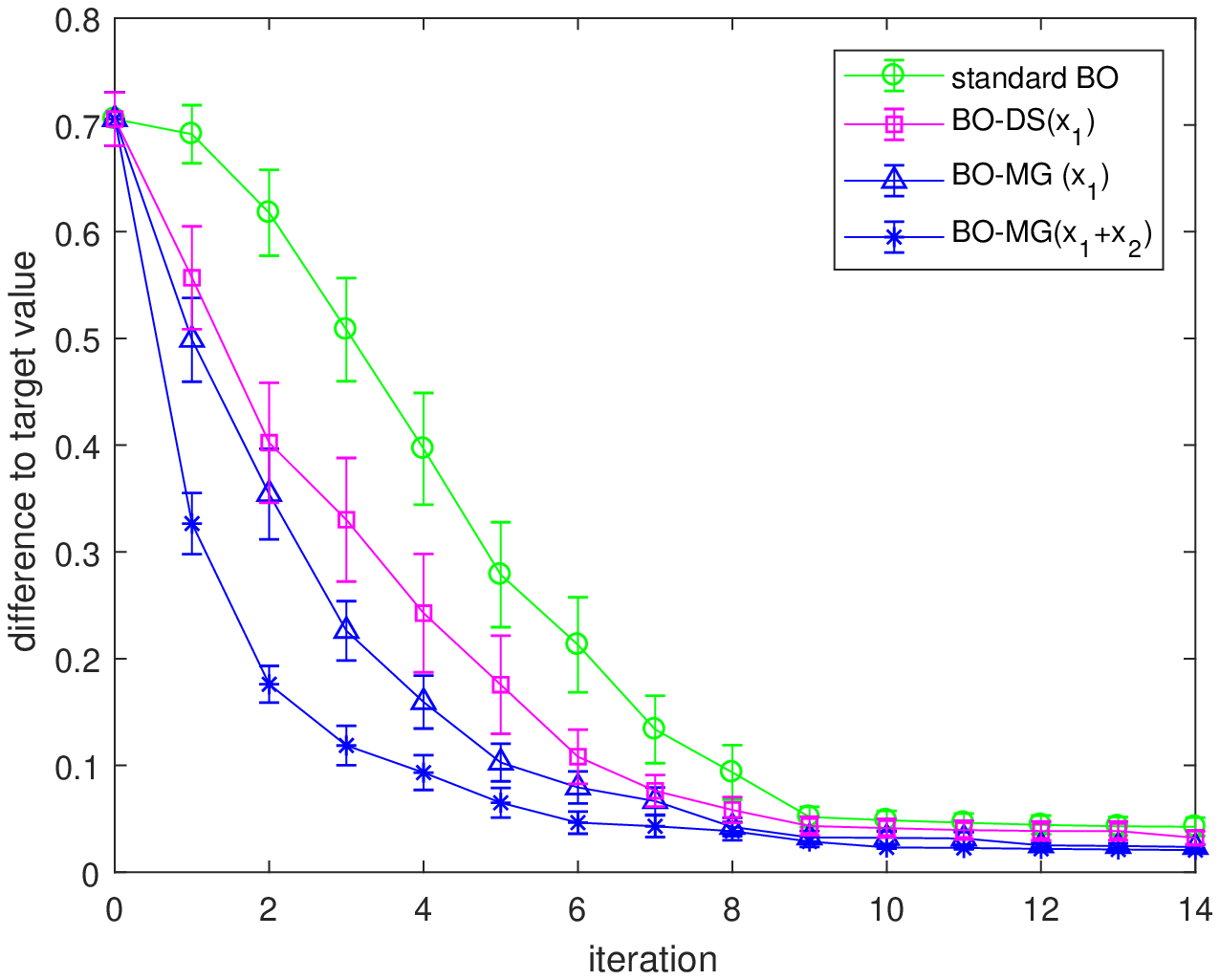}}\subfloat[7D function $f_{6}$.]{\centering{}\includegraphics[width=0.33\textwidth]{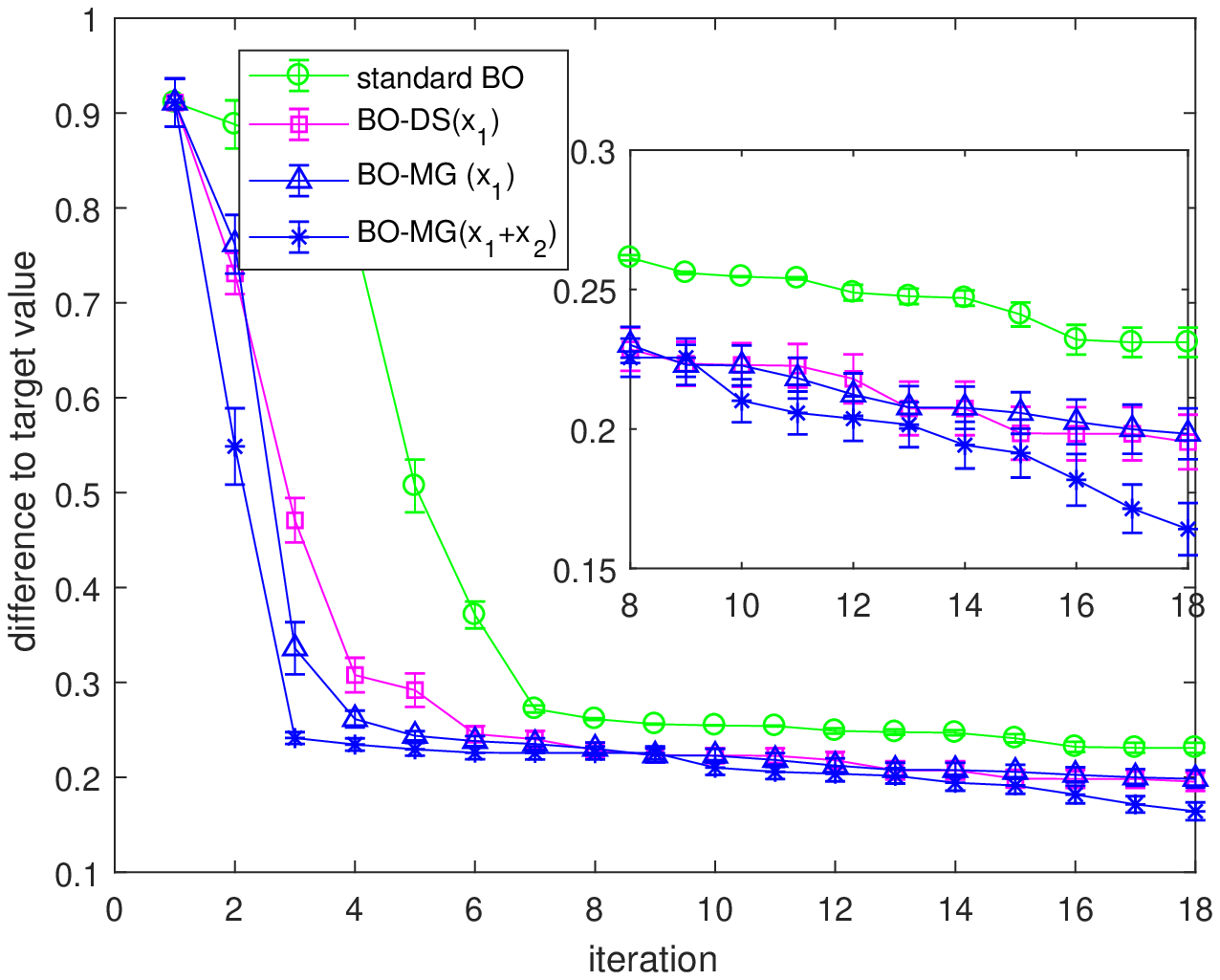}}
\par\end{centering}
\caption{\label{fig:The-simulation-results}The results of optimizing benchmark
functions. The graph shows the comparison of difference to the target
value between different algorithms. The vertical axis represents the
difference to the target value. }
\end{figure*}

(a) 2D function: $f_{1}(\boldsymbol{x})=\frac{1}{20}(x_{1}-5){}^{2}+\frac{1}{20}(x_{2}-4){}^{2}$,
$f_{T}=1.5$, $\boldsymbol{x}\in[0,5]$; 

(b) 5D function: $f_{2}(\boldsymbol{x})=\frac{1}{30}(x_{1}-3){}^{2}+\frac{1}{30}(x_{2}-2){}^{2}+\mathcal{GN}(x_{3:5}|\boldsymbol{0},\boldsymbol{1})$,
$f_{T}=1.5$, $\boldsymbol{x}\in[-2,3]$, where $\mathcal{GN}(x_{3:5}|0,1)$
is a un-normalized Gaussian PDF for $x_{3}\sim x_{5}$;

(c) 7D function: $f_{3}(\boldsymbol{x})=\frac{1}{30}(x_{1}-3){}^{2}+\frac{1}{30}(x_{2}-2){}^{2}+\mathcal{GN}(x_{3\text{:7}}|\boldsymbol{0},\boldsymbol{1})$,
$f_{T}=1.3$, $\boldsymbol{x}\in[-3,3]$, where $\mathcal{GN}(x_{3:7}|0,1)$
is a un-normalized Gaussian PDF for $x_{3}\sim x_{7}$;

(d) 2D function: $f_{4}(\boldsymbol{x})=\frac{1}{20}(x_{1}-5)x_{2}$,
$f_{T}=0.8$, $\boldsymbol{x}\in[0,5]$; 

(e) 5D function: $f_{5}(\boldsymbol{x})=\frac{1}{20}(x_{1}-5)x_{2}+\mathcal{GN}(x_{3:5}|\boldsymbol{0},\boldsymbol{1})$,
$f_{T}=1.5$, $\boldsymbol{x}\in[0,5]$,

(f) 7D function: $f_{6}(\boldsymbol{x})=\frac{1}{20}(x_{1}-5)x_{2}+\mathcal{GN}(x_{3\text{:7}}|\boldsymbol{0},\boldsymbol{1})$,
$f_{T}=1.5$, $\boldsymbol{x}\in[0,5]$, 

$f_{1}$ , $f_{2}$ and $f_{3}$ are monotonically decreasing with
$x_{1}$ at the given search space. $D+1$ initial observations are
randomly sampled. The optimization results for $f_{1}$, $f_{2}$
and $f_{3}$ are shown respectively in Figure \ref{fig:The-simulation-results}
(a), (b) and (c). We see that \emph{BO-MG} approaches the specified
target quicker than standard BO. Note that \emph{BO-DS} performs
better in the beginning than \emph{BO-MG} in the 2D function. It is
possible since derivative signs away from optimum can still make effectiveness
on the optimum on the low-dimensional space. However, it does not
happen in higher dimensions. Further, we also run the target optimization
for $f_{4}$, $f_{5}$ and $f_{6}$ which are monotonically decreasing
with $x_{1}$ and increasing with $x_{2}$ at the given search space.
Results show that \emph{BO-MG} converges faster than other baselines.

\subsection{Hyperparameter tuning in neural network}

\begin{figure}
\begin{centering}
\includegraphics[width=0.8\columnwidth]{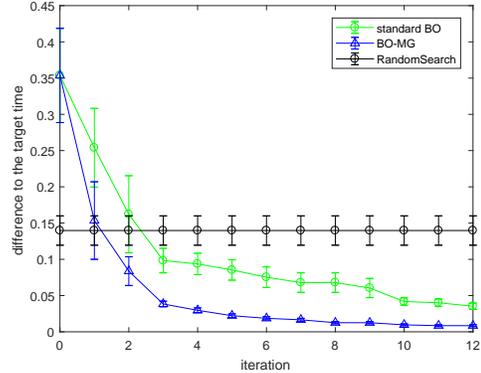}
\par\end{centering}
\caption{\label{fig:The-NN-results}Comparable performance of different algorithms
on hyperparameter tuning in neural network. The vertical axis represents
the difference to the pre-set test time. }
\end{figure}

We test our algorithm for hyperparameter tuning in neural networks.
The goal is to obtain the number of hidden neurons in each layer
for a stipulated (target) test time. We know that the test time increases
with the number of neurons i.e. it is monotonic with the number of
neurons. We split the MNIST dataset into training and testing data.
The target test time is set at 2s (A Xeon Quad-core PC 2.6 GHz with
16 GB of RAM is used). We assume that the number of neurons are the
same in each layer and allowed to vary between 10 to 1600. The other
hyperparameters in this neural network includes hidden layers (10),
dropout rate at the input layer (0.2), dropout rate at the hidden
units (0.5), learning rate for 10 layers (0.9980, 0.9954, 0.9543,
0.8902, 0.8138, 0.6519, 0.5223, 0.4184, 0.3352, 0.2685). We only optimize
the number of hidden neurons given a target test time. Result are
shown in Figure \ref{fig:The-NN-results}. \emph{BO-MG} approaches
the target time significantly quicker than standard BO and random
search. 20 out of 20 runs (100\%) in BO-MG achieve 0.05s difference
to the target test time whilst only 15 runs (75\%) in standard BO
and 6 runs (30\%) in random search reach target test time. The expected
number of neurons in BO-MG is 765 (standard deviation: 49) and that
in standard BO is 768 (standard deviation: 75). 

\subsection{Optimization of short fibers with target length}

\begin{figure*}
\begin{centering}
\subfloat[]{\begin{centering}
\includegraphics[width=0.45\textwidth]{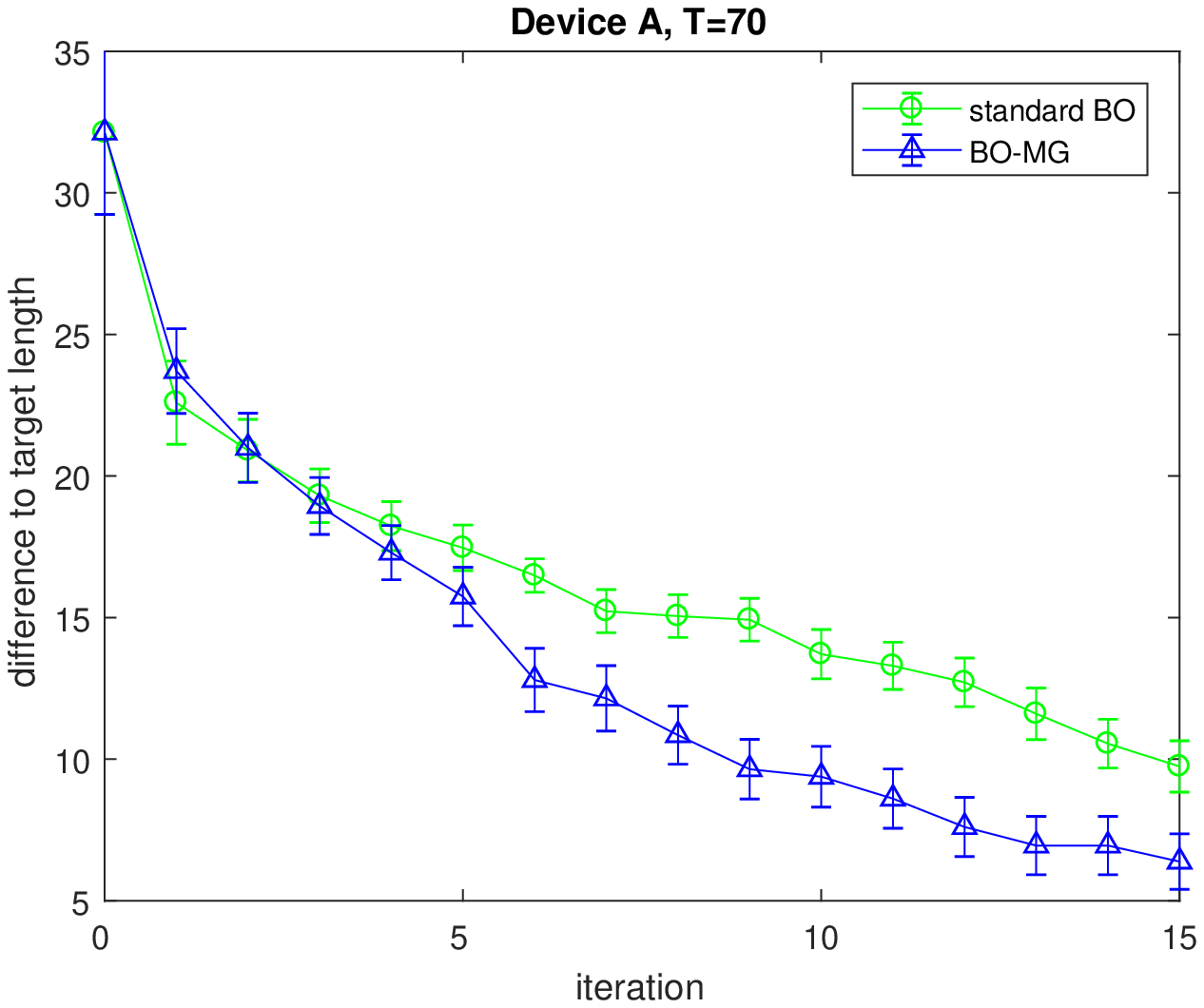}
\par\end{centering}
}\subfloat[]{\begin{centering}
\includegraphics[width=0.45\textwidth]{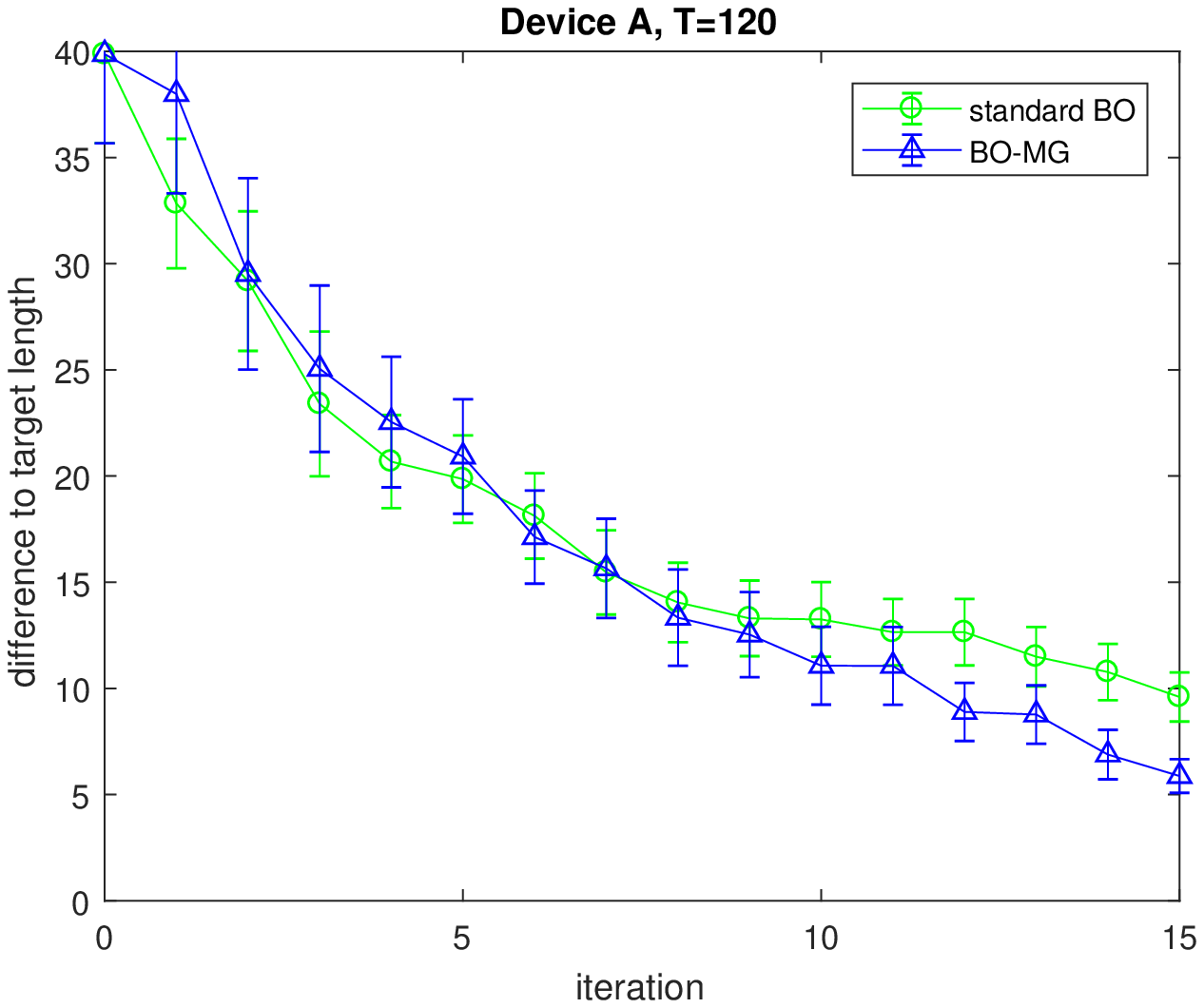}
\par\end{centering}
}
\par\end{centering}
\begin{centering}
\subfloat[]{\begin{centering}
\includegraphics[width=0.45\textwidth]{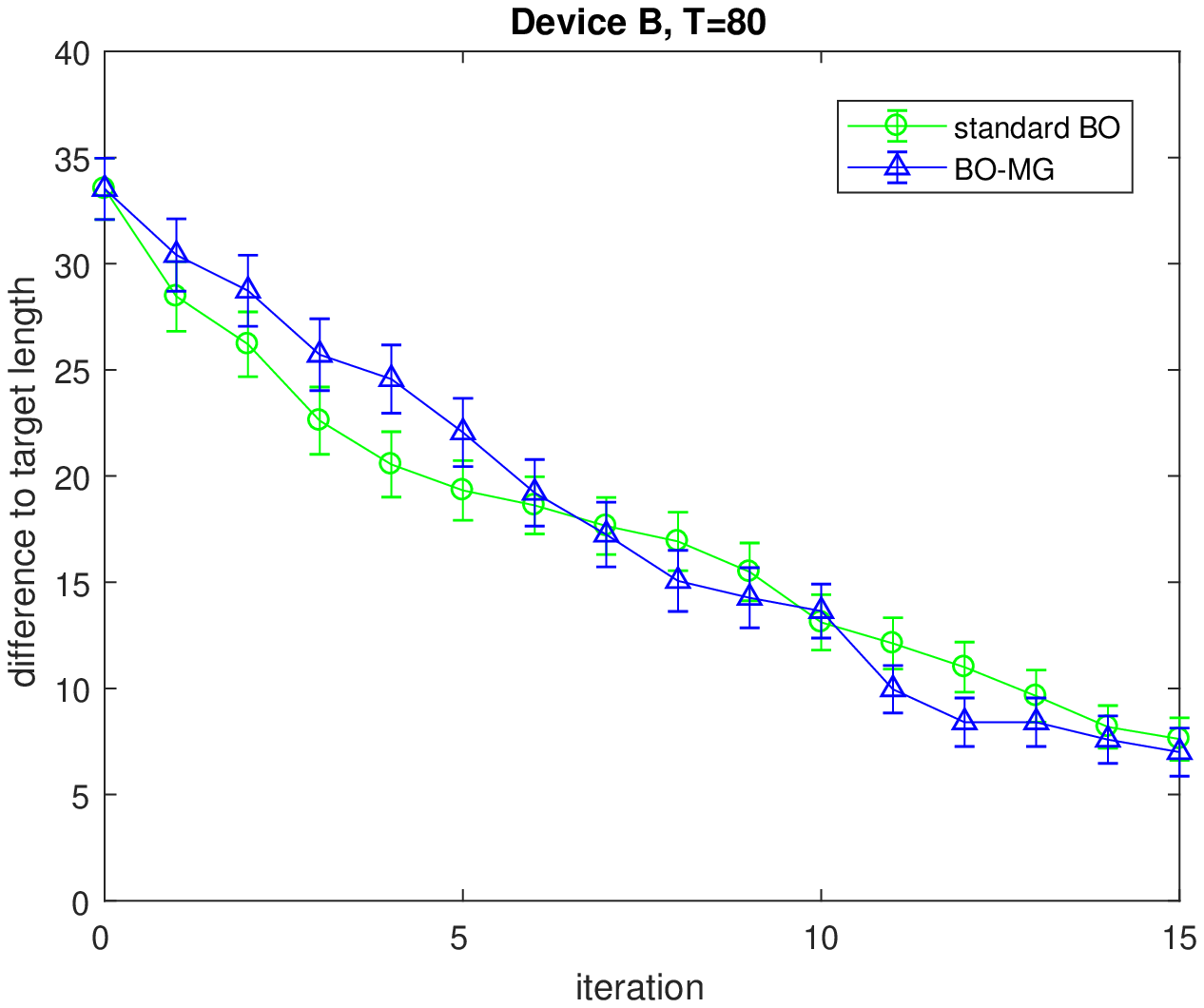}
\par\end{centering}
}\subfloat[]{\begin{centering}
\includegraphics[width=0.45\textwidth]{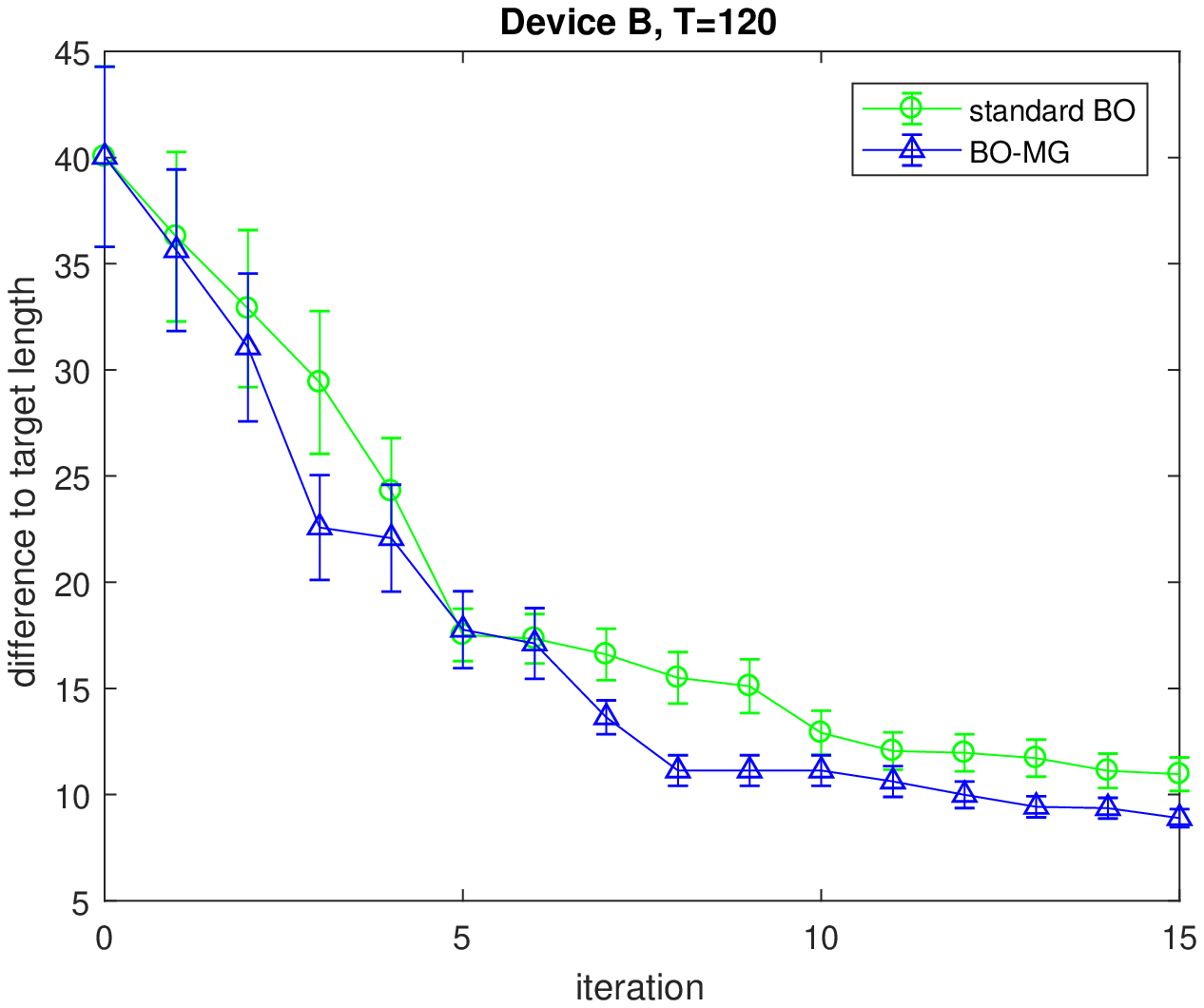}
\par\end{centering}
}
\par\end{centering}
\caption{\label{fig:SPF flowchart-1} Optimization of Short Polymer Fibre with
specified target lengths:\emph{ BO-MG} vs standard BO. The vertical
axis represents the difference from target length $(T)$. \emph{Results
for Device A} are (a)$T=70\mu m$ and (b)$T=120\mu m$; \emph{Results
for for Device B} (c)$T=80\mu m$ and (d)$T=120\mu m$\emph{.}}
\end{figure*}

We test our algorithm on a real-world application: optimizing short
polymer fiber (SPF) for a specified target length \cite{Li_rapidBO_2017}.
This involves the injection of one polymer into another in a special
microfluidic device of given geometry - Figure \ref{fig:SPF_flowchart}
before. To achieve the targeted SPF length specification, we optimize
five parameters: \emph{geometric factors}: channel width ($mm$),
constriction angle ($degree$), and device position ($mm$); and,\emph{
flow factors}: butanol speed ($cm/s$), polymer concentration ($ml/h$).
Our experimenter collaborators have confirmed that the fiber length
monotonically decreases with respect to the butanol speed. The goal
of this task is to leverage this prior knowledge to facilitate the
optimization. We test our algorithm on two\textbf{ devices}\emph{,}
and in\emph{ }each device we conduct experiments to satisify two different
targets\emph{ :}
\begin{itemize}
\item \emph{Device A }uses a gear pump \emph{\cite{sutti2014apparatus}}.
The butanol speed used is 86.42, 67.90 and 43.21. The target length
specifications are $70\mu m$ and $120\mu m$.
\item \emph{Device B} uses a lobe pump \emph{\cite{sutti2014apparatus}},
and has different plumbing configuration than device A, while retaining
the main fibre production chamber. The butanol speeds are equally
spaced: 98, 63 and 48. The target length specifications are $80\mu m$
and $120\mu m$\emph{.}
\end{itemize}
We seed the process with five random experiments. We compare \emph{BO-MG}
to standard BO in Figure \ref{fig:SPF flowchart-1} displaying the
distance to the target length at each iteration. \emph{BO-MG }approaches
the target faster than standard BO in 3 out of 4 target lengths and
performs similar in 1 out of 4 target lengths. The reduction in the
number of experiments is significant. Although we only show the difference
to target length vs iteration in the graphs, the real cost difference
is much larger. For example, in Figure \ref{fig:SPF flowchart-1}(a),
BO-MG takes 10 iterations to reach 10um difference to the target while
the standard BO takes 15 iterations. Mapping to the real time, the
standard BO takes 3 days more than BO-MG. It firmly establishes the
utility of using prior knowledge through our proposed framework. 
\begin{figure*}
\begin{centering}
\subfloat[]{\begin{centering}
\includegraphics[width=0.45\textwidth]{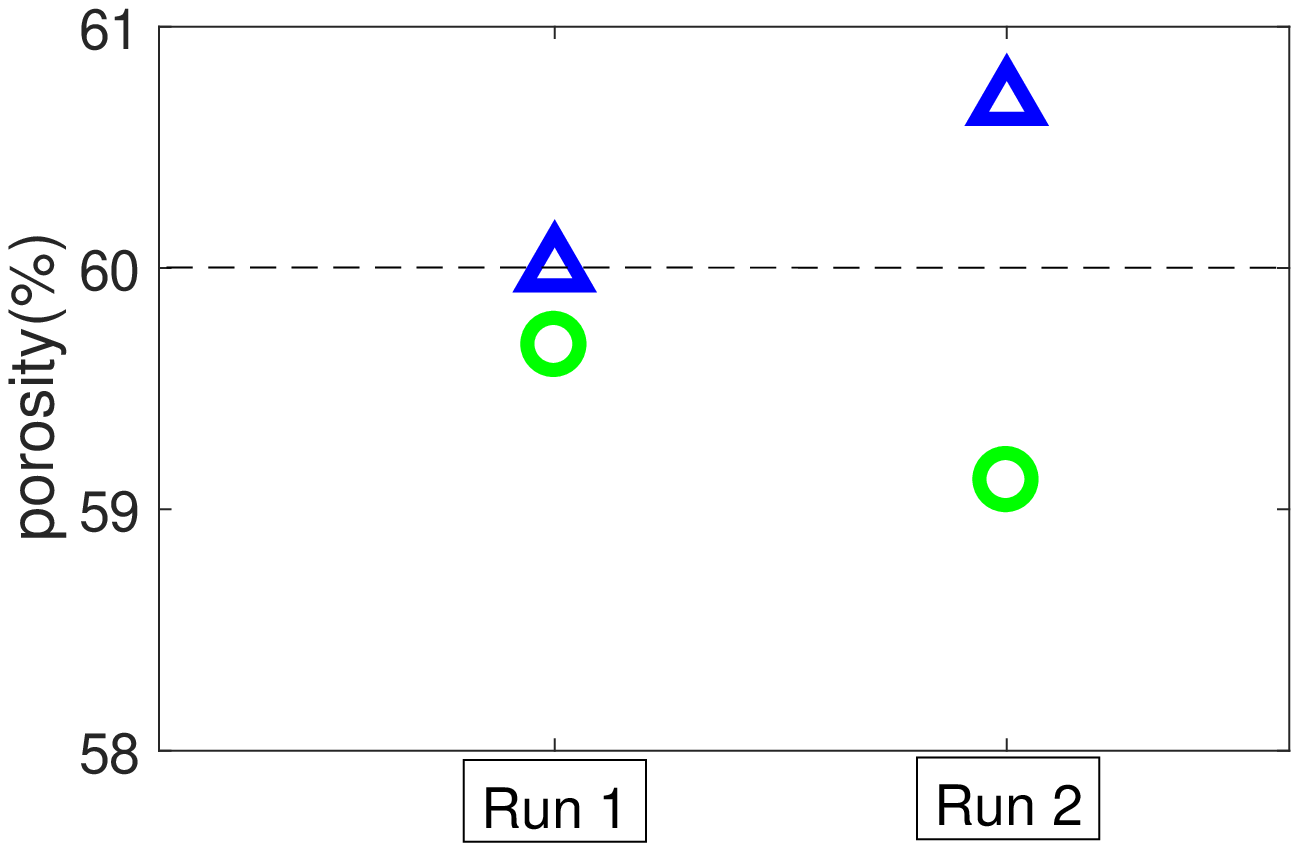}
\par\end{centering}
}\subfloat[]{\begin{centering}
\includegraphics[width=0.4\textwidth]{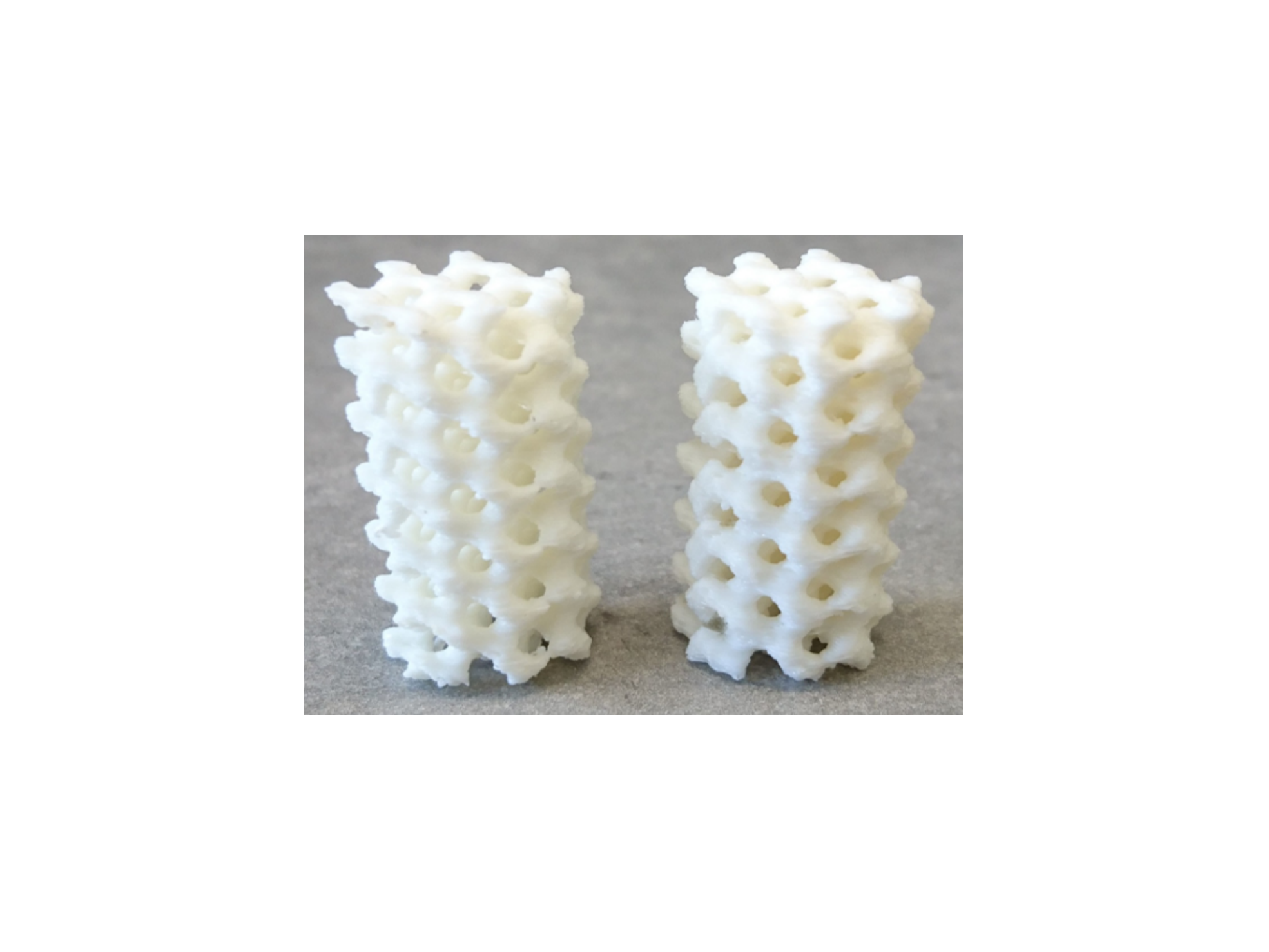}
\par\end{centering}
}
\par\end{centering}
\centering{}\caption{\label{fig:3d porosity}(a) Optimization of scaffold for a target
porosity 60\%. Standard BO (circle) \emph{vs BO-MG }(triangle) Results
for two independent runs are shown; (b) 3D printed scaffolds with
final\emph{ BO-MG} recommendations: Scaffold porosity 60\% (left)
and scaffold porosity 50.27\% (right). }
\end{figure*}

\subsection{Optimization of scaffold with target porosity in 3D printing}

With the maturity of 3D printing processes, complex three dimensional
porous architectures, or scaffolds, are becoming a favorable feature
in a range of product designs applications ranging from topology optimization
to tissue engineering structures. Such scaffold structures could not
be fabricated by any other form of technology. The ability to derived
precise solutions for the overall porosity of a resulting scaffold
can be problematic, requiring laborious trial and error based approaches
to derive a solution. 

Our objective is to derive solution for reduced material consumption
when creating a cylindrical structure, with two absolute porosity
targets of 50\% or 60\%. To adjust the porosity, the thickness of
the scaffold was adjusted by uniformly projecting the surface outward
closing the free volume. This projection was dictated by a design
software parameter, named the smallest detail, which has a lower value
of 0.05 and can be adjusted in increments of 0.001. 

We employ the \emph{BO-MG} to accelerate scaffold design to achieve
the two targeted porosities with fewer number of experiments. We have
a hunch that the porosity decreases with the smallest detail. Starting
from three random points, we recommended three sequential experiments
for targeted porosity 60\%. The search range of the smallest detail
is between 0.05 and 2. We run this process independently twice and
compare the best suggested one from different algorithms. The result
for T=60\% is shown in Figure \ref{fig:3d porosity}. \emph{BO-MG}
recommendations are closer to the targeted porosity. 

We also exploit all previous experimental results to suggest only
one experiment for targeted porosity 50\%. The recommended experiment
from \emph{BO-MG} acheives porosity of 50.27\% whilst standard BO
 reaches porosity of 49.22\%. The results clearly demonstrate the
effectiveness of our method.

\section{Conclusion\label{sec:conclusion}}

We have proposed a Bayesian optimization algorithm to incorporate
the hunches experimenters possess about the change of experimental
results with respect to certain variables to accelerate experimental
designs. We have explicitly discussed the monotonicity information
and how to model it into Bayesian optimization framework. We also
provide the regret bound for our method to demonstrate its convergence.
The experimental results show that the proposed algorithm significantly
outperforms the standard Bayesian optimization and it reduces significant
cost in real world applications. Regarding the future work we seek
a smart way to automatically detect the trends of the function so
that BO strategies can switch freely between different trends. More
broadly we have envisaged the benefit of the use of monotonicity information
in Bayesian optimization and exploring the use of other types of prior
knowledge is a promising direction for efficient experimental design.
\\
\textbf{Acknowledgment:} This research was partially funded by the Australian Government through the Australian Research Council (ARC). Prof Venkatesh is the recipient of an ARC Australian Laureate Fellowship (FL170100006)

\bibliographystyle{IEEEtran}

\end{document}